%% file: compiler_paper.tex
\documentclass[journal]{IEEEtran}

\usepackage[english]{babel}

\usepackage{ifpdf}

\usepackage{cite} 
\usepackage{url}
\usepackage{hyperref}

\ifCLASSINFOpdf
	\usepackage[pdftex]{graphicx}
	\graphicspath{{./figures/}}
\else
	\usepackage[dvips]{graphicx}
	\graphicspath{./figures/}
\fi
\usepackage{color}
\usepackage{pgf, tikz, pgfplots}
\usetikzlibrary{shapes, arrows, automata, plotmarks}
\usetikzlibrary{calc,hobby,decorations}
\usepgfplotslibrary{fillbetween}
\usepackage{soul}

\usepackage[cmex10]{amsmath}
\usepackage{amsfonts, amssymb, amsthm}
\usepackage{mathrsfs}

\usepackage{algorithm,algpseudocode}
	\algnewcommand{\LeftComment}[1]{\Statex \(\triangleright\) #1}
\newcommand{\makeboxlabel}[1]{#1\hfill}

\usepackage{caption}
\usepackage{multirow}
\usepackage{rotating}
\usepackage{subcaption}
\input{./latex_resources/mySections.tex}

\input{./latex_resources/mySymbol.sty}
\input{./latex_resources/pennColors.sty}

\DeclareMathAlphabet\mathrsfso {U}{rsfso}{m}{n}
\newcommand{\End}[1]{\text{End}(\mathcal{#1})}
\newcommand{\vertiii}[1]{{\left\vert\kern-0.25ex\left\vert\kern-0.25ex\left\vert #1 
		\right\vert\kern-0.25ex\right\vert\kern-0.25ex\right\vert}}

\newtheorem{example}{\hspace{0pt}\bf Example}

\newtheorem{theorem}{\hspace{0pt}\bf Theorem}
\newtheorem{corollary}{\hspace{0pt}\bf Corollary}

\newtheorem{remark}{\hspace{0pt}\bf Remark}

\newtheorem{definition}{\hspace{0pt}\bf Definition}

\begin{document}


\title{Convolutional Filters and Neural Networks with \\ Non Commutative Algebras}
\author{Alejandro~Parada-Mayorga, 
	        Landon Butler
            and Alejandro Ribeiro
\thanks{Dept of Electrical and Systems Engineering, University of Pennsylvania. Email: alejopm@seas.upenn.edu, landonb3@seas.upenn.edu, aribeiro@seas.upenn.edu. Supported by NSF-Simons MoDL Award \#2031985 
}}

\markboth{IEEE Transactions on Signal Processing (submitted)}%
{Shell \MakeLowercase{\textit{et. al.}}: Bare Demo of IEEEtran.cls for Journals}
\maketitle


\input{\pathsections/sec_abstract.tex}



\IEEEpeerreviewmaketitle


\input{\pathsections/sec_introduction.tex}


\input{\pathsections/sec_RTASM.tex}


\input{\pathsections/sec_Spectral_Representation_Filters.tex}


\input{\pathsections/sec_algNN.tex}


\input{\pathsections/perturbations_and_stability.tex}


\input{\pathsections/sec_numexp.tex}


\input{\pathsections/sec_discussion.tex}


\appendices


\input{\pathsections/sec_basic_algebra.tex}


\input{\pathsections/sec_group_sp_nn.tex}


\input{\pathsections/sec_quaternion_sp_nn.tex}


\input{\pathsections/sec_proofTheorems.tex}


\bibliography{bibliography}
\bibliographystyle{unsrt}

\clearpage



\appendices
\input{\pathsections/appendix_HsFrechet.tex}

\ifCLASSOPTIONcaptionsoff
  \newpage
\fi

\end{document}

%% file: latex_resources/mySections.tex
\usepackage{needspace}



%% file: v25/sec_abstract.tex

\begin{abstract}

In this paper we introduce and study the algebraic generalization of non commutative convolutional neural networks. We leverage the theory of algebraic signal processing to model convolutional non commutative architectures, and we derive concrete stability bounds that extend those obtained in the literature for commutative convolutional neural networks. We show that non commutative convolutional architectures can be stable to deformations on the space of operators. We develop the spectral representation of non commutative signal models to show that non commutative filters process Fourier components independently of each other. In particular we prove that although the spectral decompositions of signals in non commutative models are associated to eigenspaces of dimension larger than one, there exists a trade-off between stability and selectivity, which is controlled by matrix polynomial functions in spaces of matrices of low dimension. This tradeoff shows how when the filters in the algebra are restricted to be stable, there is a loss in discriminability that is compensated in the network by the pointwise nonlinearities. The results derived in this paper have direct applications and implications in non commutative convolutional architectures such as group neural networks, multigraph neural networks, and quaternion neural networks, for which we provide a set of numerical experiments showing their behavior when perturbations are present. 

\end{abstract}

%% file: v25/sec_introduction.tex


\section{Introduction}


Deep learning relies on parameterizations given by the composition of layers which are themselves compositions of linear operators with pointwise nonlinearities. In problems that involve high dimensional inputs it becomes necessary to exploit their structure to reduce the complexity of the learning parametrization. This is often accomplished with the use of particular instantiations of \emph{convolutional} filter banks. The most notable examples of this approach are the use of standard (Euclidean) \emph{convolutional} filters for learning with time signals and images, graph \emph{convolutional} filters for learning with graphs and graph signals, and group \emph{convolutions} for signals with group symmetries. The success of convolutional neural networks of different types -- Euclidean, graph, and group -- is in part due to how convolutions leverage symmetries of the domain and the signals. However, both filter banks and convolutional neural networks are equally good at leveraging these symmetries~\cite{parada_algnn,gama2020stability,gamabruna_diffscattongraphs}. Yet, it is the latter that are more successful at learning. This empirical observation prompts a search for properties that explain the better performance of convolutional neural networks relative to the corresponding convolutional filter banks.

In the case of Euclidean convolutions and graph convolutions part of the insight into the relative performance of convolutional filter banks and neural networks follows from their respective responses to deformations~\cite{mallat_ginvscatt, gama2020stability} \black{-- deformations in the domain of the signals in~\cite{mallat_ginvscatt} and deformations on the space of graph matrix representations in~\cite{gama2020stability}}. These works prove that stability to deformations requires filters that do not distinguish different high frequency components. It follows that convolutional neural networks can better trade-off discriminability and stability as the pointwise nonlinearity mixes frequency components across layers. Recent analysis conducted on the operator space has demonstrated that the shared stability properties of Euclidean and graph convolutions are a consequence of their shared algebraic structure~\cite{parada_algnn}. It then follows that whenever we use convolutional parameterizations in machine learning we can expect convolutional neural networks with multiple layers to outperform convolutional filter banks. These results build on the theory of algebraic signal processing (ASP) which provides a common language for describing convolutional filters of different types~\cite{algSP0}. They apply to Euclidean, graph, and group convolutions as well as to a large class of less ubiquitous convolutional filters and neural networks~\cite{parada_algnn, parada_algnnconf}.

Although these results provide valuable insights, they are limited to commutative operators. Namely, to signal processing architectures in which filters commute. This is not true in general as non commutative signal processing arises naturally in multiple scenarios~\cite{Knyazev2018SpectralMN, Weiler2019GeneralES, Weiler20183DSC, Worrall2017HarmonicND, Finzi2020GeneralizingCN, Finzi2021APM, Hoffmann2020AlgebraNets, gaudet2018deep,Kumar2022AlgebraicCF}. For instance, multigraphs arise when nodes are related by several different types of edges~\cite{Knyazev2018SpectralMN,Butler2022ConvolutionalLO}. In this case it is natural to define multigraph filters by combining diffusions across matrix representations of the graphs defined by each individual type (Example \ref{ex_multgsp}). The resulting filters do not commute except in the rare eventuality that the different matrix representations themselves commute. Filters that do not commute also arise with signal models on groups like $\mbE (n)$, or $SO(n)$~\cite{Weiler2019GeneralES, Weiler20183DSC, Worrall2017HarmonicND,Bronstein2021GeometricDL}, or any nontrivial Lie group~\cite{Finzi2020GeneralizingCN, Finzi2021APM, Bronstein2021GeometricDL,weiler2021coordinate}. Non commutative filters appear here because filters are associated with group symmetries and it is common for groups to have symmetries that do not commute -- such as rotations in $SO(3)$.

Existing stability results do not apply to any of these non commutative settings but empirical results show the usual advantage of layered neural networks with respect to filter banks~\cite{Butler2022ConvolutionalLO,Bronstein2021GeometricDL,Finzi2020GeneralizingCN, Finzi2021APM,Kumar2022AlgebraicCF}. Such an empirical observation poses the question of whether similar stability results hold for non commutative convolutional architectures. That is, whether convolutional versions of architectures such as multigraph neural networks~\cite{Butler2022ConvolutionalLO,msp_icassp2023}, neural networks on groups~\cite{Bronstein2021GeometricDL,Kumar2022AlgebraicCF}, quaternion (graph) neural networks~\cite{gaudet2018deep,Nguyen2020QuaternionGN,Parcollet2018QuaternionRN,Qiu2020QuaternionNN,xia2018echo,parcollet2020survey,grassucci2022quaternion}, hypercomplex algebras neural networks~\cite{Grassucci2021PHNNsLN}, hyperbolic neural networks~\cite{857886,nitta2008decision}, octonion neural networks~\cite{popa2016octonion}, and Clifford algebras neural networks~\cite{bayro2005geometric,buchholz2008clifford,buchholz2001clifford} can be stable to deformations and, more to the point, whether stability requires spectral restrictions analogous to those of Euclidean, graph, and commutative group neural networks. The goal of this paper is to answer this question in the affirmative. To do so our first two contributions are the following:
\smallskip
\begin{list}
      {}
      {\setlength{\labelwidth}{26pt}
       \setlength{\labelsep}{-3pt}
       \setlength{\itemsep}{10pt}
       \setlength{\leftmargin}{26pt}
       \setlength{\rightmargin}{0pt}
       \setlength{\itemindent}{0pt} 
       \let\makelabel=\makeboxlabel
       }
\item [(C1)] We define convolutional filters (Section~\ref{sec_algsigmod}) and neural networks (Section~\ref{sec_Algebraic_NNs}) with non commutative algebras. 
\item [(C2)] We develop the spectral representation of non commutative signal models (Section \ref{sec:spectopt}).  %
\end{list}
\medskip
Non commutative filters are just a particular case of algebraic filters \cite{algSP0} and the construction of algebraic neural networks with non commutative filters is a close to verbatim extension of algebraic neural networks \cite{parada_algnn, parada_algnnconf}. The development of spectral representations, however, has substantial differences. Commutative signal models define Fourier decompositions as projections on single dimensional subspaces and, consequently, frequencies are defined as scalars. In non commutative signal models Fourier decompositions are projections in multidimensional subspaces and frequencies are, consequently, matrices of corresponding dimensions (Definition~\ref{def:foudecomp}). Despite this significant difference we can still prove the equivalent of an spectral representation theorem of non commutative algebraic filters:
\smallskip
\begin{list}
      {}
      {\setlength{\labelwidth}{26pt}
       \setlength{\labelsep}{-3pt}
       \setlength{\itemsep}{10pt}
       \setlength{\leftmargin}{26pt}
       \setlength{\rightmargin}{0pt}
       \setlength{\itemindent}{0pt} 
       \let\makelabel=\makeboxlabel
       }
\item [(C3)] Non commutative algebraic filters process Fourier components independently of each other (Theorem \ref{thm_filtspec}). 
\end{list}
\smallskip
Contribution (C3) is the equivalent of the claim that commutative algebraic filters process frequency components independently of each other. As in the case of commutative filters, Contribution (C3) implies that non commutative algebraic filters are completely characterized by their frequency responses. The difference is that in non commutative filters the frequency response is a \emph{matrix} polynomial (Definition \ref{definition:filtspecrep}). This is in contrast to the scalar polynomials that define frequency representations in commutative signal models and a consequence of the fact that frequencies are matrices, not scalars.

It is this difference between frequency representations of commutative and non commutative filters -- matrix versus scalar polynomials -- that prevents the results derived in~\cite{parada_algnn} to be applied to non commutative convolutional architectures. The main technical contribution of this paper is to generalize the analysis of \cite{parada_algnn} to show that analogous results hold for non commutative algebraic signal models. In particular, we prove that: 
\smallskip
\begin{list}
      {}
      {\setlength{\labelwidth}{26pt}
       \setlength{\labelsep}{-3pt}
       \setlength{\itemsep}{10pt}
       \setlength{\leftmargin}{26pt}
       \setlength{\rightmargin}{0pt}
       \setlength{\itemindent}{0pt} 
       \let\makelabel=\makeboxlabel
       }
\item [(C4)] Non commutative algebraic convolutional filters \emph{can} be stable to additive and multiplicative perturbations of the algebraic signal model (Section~\ref{sec:stabilitytheorems}). 
\item [(C5)] Non commutative algebraic neural networks inherit the stability properties of algebraic convolutional filters (Section~\ref{sec_stability_AlgNN}). 
\end{list}
\smallskip
The proof of Contribution (C4) and (C5) requires that filters have two spectral properties (Definition~\ref{def_lipschitz_intlipschitz}). The first property is a generalized Lipschitz condition in which changes in the filter's frequency response are upper bounded by a linear function of the frequency's norm. The second condition is that the Fr\'echet derivative of the filter's response acting on a frequency matrix has bounded norm. These two conditions limit the discriminability of non commutative filters. They respectively imply that: (i) The variability of a stable filter must be bounded by the norm of the difference between frequency matrices. (ii) The variability of stable filters must decrease with the norm of the frequency matrix. Stability of commutative filters requires analogous conditions on scalar frequencies. Since imposing conditions on frequency matrices is more stringent, our stability results suggest that non commutative filters require more stringent filter restrictions to attain the same level of stability. 

To illustrate our theoretical results we provide a set of numerical experiments for non commutative convolutional multigraph and quaternion neural networks (Section~\ref{sec_numexp}).

%% file: v25/sec_RTASM.tex

%

\section{Non Commutative Filters} \label{sec_algsigmod}


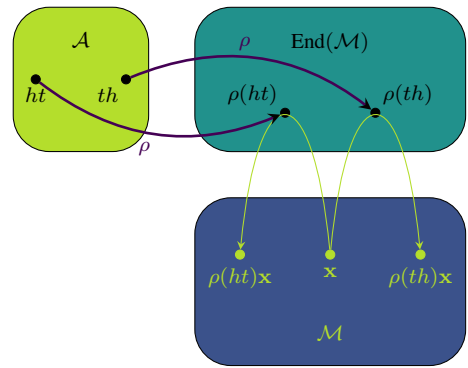
\begin{figure}
    \centering
    \input{./figures/50_asp}
    \caption{Non commutative algebraic signal model. The algebraic filters $th$ and $ht$ are \textit{realized} physically in $\End{M}$ to process the signals $\bbx$ which are modeled as elements of $\ccalM$.}
    \label{fig:my_ncASP}
\end{figure}


In this section we introduce the generalization of non commutative signal models on arbitrary domains under the lens of algebraic signal processing (ASP). An algebraic signal model (ASM) is defined as the triplet
\begin{equation}\label{eqn_ASP_signal_model}
   (\ccalA,\ccalM,\rho),
\end{equation}   
where $\ccalA$ is a unital associative algebra, $\ccalM$ is a vector space, and $\rho:\ccalA\rightarrow\text{End}(\ccalM)$ is a homomorphism between the algebra $\ccalA$ and the set of endomorphisms in the vector space $\ccalM$~\cite{algSP0}. We recall that an algebra is a vector space with a closed operation of product, the endomorphims of $\ccalM$ are the set of linear maps from $\ccalM$ onto itself and a homomorphism is a linear map between algebras that preserves the product operation. We refer the reader to Appendix~\ref{sec_basic_algebra} for a review of these concepts and the discussion of some examples as well.

For our discussion we consider that $\ccalA$ and $\ccalM$ as vector spaces are defined on an \textit{algebraically closed field} $\mbF$ which for the sake of simplicity will be considered $\mbF =\mbC$. However, we remark that the material discussed is valid for any algebraically closed Field. In the triplet $(\ccalA,\ccalM,\rho)$ the pair $(\ccalM,\rho)$ constitutes a \textit{representation} of $\ccalA$ in the context of \textit{representation theory of algebras}. 





%

The processing of information in the context of~(\ref{eqn_ASP_signal_model}) is known as algebraic signal processing (ASP). The signals are modeled as elements of $\ccalM$, the filters are defined as elements of $\ccalA$ and their realization is given by $\rho$. The rules or laws governing the operations are given by the structure of $\ccalA$ and $\rho$ translates those operations into actions on the elements of $\ccalM$. Then, the filtering of a signal $\bbx\in\ccalM$ by an algebraic filter $a\in\ccalA$ is given by
\begin{equation}\label{eqn_ASP_filter_outputs}
   \bby = \rho(a) \bbx.
\end{equation}   
In \eqref{eqn_ASP_filter_outputs} we have a generalized representation of the convolution operation between a filter and a signal. Particular instantiations of \eqref{eqn_ASP_filter_outputs} lead to the traditional signal processing models of time signals and images and more sophisticated models such as graph signal processing (GSP), graphon signal processing (WSP) among others~\cite{parada_algnn, parada_algnnconf, parada_quiversp}.

Since it is $\ccalA$ the algebraic object defining the structural rules of information processing, the non commutativity in an algebraic model is associated to $\ccalA$. This is, any non commutative convolutional signal model is given by $(\ccalA,\ccalM,\rho)$ where $\ccalA$ is a non commutative algebra. For the reader unfamiliar with algebraic concepts, we remark that one representative example of this type of algebra is $M_{n\times n}(\mbC)$, which is the set of matrices of dimension $n\times n$ with entries in $\mbC$. It is an algebra, since it is a vector space over $\mbC$ and the product of two matrices of size $n\times n$ is again a matrix of the same size.

Now we recall the notion of generators in algebraic signal models.

%
\begin{definition}[Generators] \label{def_generators} For an associative algebra with unity $\ccalA$ we say the set $\ccalG\subseteq\ccalA$ generates $\ccalA$ if all $a\in\ccalA$ can be represented as polynomial functions of the elements of $\ccalG$. We say elements $g\in\ccalG$ are generators of $\ccalA$ and we denote as $a=p(\ccalG)$ the polynomial that generates $a$.
\end{definition}

%
%

Representing the elements of $\ccalA$ as polynomial functions of a set of generators highlights the fact that such generators characterize the algebra, and can be conceived as a measure of the degrees of freedom associated to the algebra. Consequently, the realization of the generators by means of the homomorphism $\rho$, called shift operators, play a central role when analyzing algebraic signal models. We introduce its formal definition next.


\begin{definition}[Shift Operators]\label{def_shift_operators} 
Let $(\ccalA, \ccalM,\rho)$ be an algebraic signal model. Then, if $\ccalG\subset\ccalA$ is a generator set of $\ccalA$, the operators $\bbS = \rho(g)$ with $g\in\ccalG$ are called shift operators. The set of all shift operators is denoted by $\ccalS$.
\end{definition}



Then, we can express any element $\rho(a)\in\text{End}(\ccalM)$ as a polynomial function of the shift operators. In particular, if $a=p(\ccalG)$ we have that
\begin{equation}\label{eqn_filter_representation}
   \rho(a) = p_\ccalM \big(\rho(\ccalG)\big) 
           = p_\ccalM\big(\ccalS\big)
           = p\big(\ccalS\big),
\end{equation}
where $p_\ccalM$ indicates the polynomial of form $p$ but whose independent variables are the shift operators. For the sake of simplicity we drop the subindex in $p$ in our subsequent discussion since it is clear where $p$ is defined from the independent variable.

To facilitate the understanding of these basic concepts we present an example.


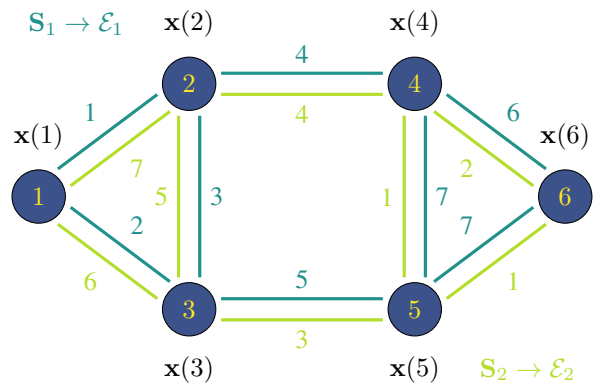
\begin{figure}
	\centering
        \input{./figures/fig13_tikz_source.tex}
	\caption{Example of a multigraph signal processing model. The signal $\bbx\in\mbR^6$ defined on the multigraph $G=(\ccalV, \{ \ccalE_1, \ccalE_2 \})$ is associated to the nodes $\ccalV$, while the shift operator $\bbS_i$ is associated to the set of edges $\mathcal{E}_i$. }
	\label{fig_multigraph}
\end{figure}



\begin{example}[Multigraph signal processing]\label{ex_multgsp}\normalfont

A multigraph consists of a common set of vertices and multiple separate sets of edges. Consider the specific case $G=\left( \ccalV, \{ \ccalE_1, \ccalE_2\} \right)$ consisting of a set of $N$ vertices $\ccalV$ and two separate edge sets $\ccalE_1$ and $\ccalE_2$ -- see Fig.~\ref{fig_multigraph}. Associated with each edge set we consider matrix representations $\bbS_1$ and $\bbS_2$. This is an algebraic signal processing model in which the vector space of signals $\ccalM=\mbC^{N}$ contains vectors in $\mbC^{N}$ with entries associated with each node of the multigraph and the algebra $\ccalA = \mbC[t_1,t_2]$ is the set of non commutative polynomials of two variables. 
The algebra $ \mbC[t_1,t_2] $ is generated by the monomials $t_1$ and $t_2$. The homomorphism $\rho$ is defined by mapping the generators $t_1$ and $t_2$ to $\rho(t_1)=\bbS_1$ and $\rho(t_2)=\bbS_2$, where the shift operators $\bbS_i$ are the matrix representations of the corresponding set of edges $\ccalE_i$. Then, if $p(t_1,t_2) = t_{1}^2 + t_1 t_2 + 2 t_2 t_1 + t_{2}^{2} +1$ the filtering in~\eqref{eqn_ASP_filter_outputs} takes the form
	\begin{multline}
	\rho \left(
	      t_{1}^2 + t_1 t_2 + 2 t_2 t_1 + t_{2}^{2} +1
	     \right) 
	\bbx
	\\
	=
	\left(
	      \bbS_{1}^2 + \bbS_1 \bbS_2 + 2 \bbS_2 \bbS_1 + \bbS_{2}^{2} +1
	\right)\bbx.   
	\end{multline}
\end{example}


We refer the reader to Appendix~\ref{ex_mult_filt_extra} in the supplementary material for a more sophisticated example on multigraph signal processing in the light of ASP. In Appendix~\ref{sec_grsp_discussion} and Appendix~\ref{sec_quaternions_sp_nn} of the paper it is shown how classical group signal processing and convolutional quaternion signal processing can be seen as particular cases of algebraic signal models.


\begin{remark}\normalfont
	It is important to highlight that the definition of an algebraic signal model is independent of any norm that could be associated to $\ccalM$, if any. As we will discuss in subsequent sections (see Section~\ref{sec:stabilitytheorems}) we will endow $\ccalM$ with a norm in order to analyze the size of perturbations, and doing this will not affect the structure of the triplet $(\ccalA,\ccalM,\rho)$. We also remark that the algebra $\ccalA$ is not associated to a norm, but it happens that $\ccalA$ can be isomorphic to algebras that are naturally endowed with a norm. This is something we exploit in Section~\ref{sec:spectopt} to characterize subsets of the algebra and spectral representations.
\end{remark}


%% file: figures/50_asp.tex

\def \scale { 1.2}
\def \unit  { \scale cm}


\definecolor{my_cp_col1}{RGB}{253, 231, 37}
\definecolor{my_cp_col2}{RGB}{180, 222,44}
\definecolor{my_cp_col3}{RGB}{94, 201, 98}
\definecolor{my_cp_col4}{RGB}{33, 145, 140}
\definecolor{my_cp_col5}{RGB}{59, 82, 139}
\definecolor{my_cp_col6}{RGB}{68, 1, 84}


\tikzstyle{set} = [ rectangle,
                    rounded corners = 0.4*\unit,
                    inner sep=0pt,
                    draw,
                    anchor = center ]

\tikzstyle{vector space} = [ set,
                             fill=my_cp_col5,
                             minimum width  = 3*\unit,
                             minimum height = 1.86*\unit]

\tikzstyle{endomorphisms} = [ vector space,
                              fill=my_cp_col4,
                              minimum height = 1.6*\unit]

\tikzstyle{algebra} = [ endomorphisms,
                        fill=my_cp_col2,
                        minimum width = 1.5*\unit]

\tikzstyle{dot} = [ circle,
                    minimum width  = 0.1*\unit,
                    fill=black,
                    inner sep=0pt,
                    draw,
                    anchor = center ]

{\fontsize{8}{8}\selectfont

\begin{tikzpicture}[-stealth, draw = black!99, scale = \scale]


   \path (0,0) node [vector space] (M) {};
   \path (M.south) ++ (0, 0.2) node [above] {\textcolor{my_cp_col2}{$\ccalM$}};
   
   \path (M) ++ (0, +0.3) node [dot,color=my_cp_col2] (x) {};
   \path (x.south) node [below] {\textcolor{my_cp_col2}{$\bbx$}};             
   \path (M) ++ (+1, +0.3) node [dot,color=my_cp_col2] (ex) {};
   \path (ex.south) node [below] {\textcolor{my_cp_col2}{$\rho(th)\bbx$}};
   
   \path (M) ++ (-1, +0.3) node [dot,color=my_cp_col2] (rex) {};
   \path (rex.south) node [below] {\textcolor{my_cp_col2}{$\rho(ht)\bbx$}};

   \path (M.north) ++ (0, 0.5) 
         node [endomorphisms, anchor=south] (End) {};
   \path (End.north) ++ (0.0, -0.2) node [below] {$\text{End}(\ccalM)$};   

   \path (End.center) ++ (0.5, -0.37) node [dot] (e) {};      
   \path (e) node [above right] {$\rho(th)$};  
   
   \path (End.center) ++ (-0.5, -0.37) node [dot] (re) {};      
   \path (re) node [above left] {$\rho(ht)$}; 
   
   \path (End.center) + (0.2,0.1) coordinate (c1);
   \path (End.center) + (0.8,0.1) coordinate (c2);   
   \path [draw, -stealth,color=my_cp_col2, thin] (x) .. controls (c1) and (c2) .. (ex);   

   \path (End.center) + (-0.2,0.1) coordinate (c1);
   \path (End.center) + (-0.8,0.1) coordinate (c2);   
   \path [draw, -stealth,color=my_cp_col2, thin] (x) .. controls (c1) and (c2) .. (rex);

   \path (End.north west) ++ (-0.5, 0) 
         node [algebra, anchor = north east] (A) {};   
   \path (A.north) ++ (0,-0.2) node [below] {$\ccalA$};   
   
   \path (A.center) ++ (+0.5,0) node [dot] (a) {};      
   \path (a) node [below left] {$th$};    
   
   \path (A.center) ++ (-0.5,0) node [dot] (ra) {};      
   \path (ra) node [below ] {$ht$};

   \path (End.west) + (-0.1,0.25) coordinate (c1);
   \path (End.east) + (+0.1,0.25) coordinate (c2);   
   \path [draw, -stealth, line width = 1.0, my_cp_col6] 
         (a) edge [bend left] node [above] {$\rho~~~$} (e);  
         
   \path (End.west) + (-0.1,0.25) coordinate (c1);
   \path (End.east) + (+0.1,0.25) coordinate (c2);   
   \path [draw, -stealth, line width = 1.0, my_cp_col6] 
         (ra) edge [bend right] node [below] {$\rho~~~$} (re);

\end{tikzpicture}

}

%% file: figures/fig13_tikz_source.tex


\definecolor{my_cp_col1}{RGB}{253, 231, 37}
\definecolor{my_cp_col2}{RGB}{180, 222,44}
\definecolor{my_cp_col3}{RGB}{94, 201, 98}
\definecolor{my_cp_col4}{RGB}{33, 145, 140}
\definecolor{my_cp_col5}{RGB}{59, 82, 139}
\definecolor{my_cp_col6}{RGB}{68, 1, 84}

\newcommand\DoubleLine[7][4pt]{%
    \path(#2)--(#3)coordinate[at start](h1)coordinate[at end](h2);
    \draw[#4]($(h1)!#1!90:(h2)$)-- node [auto=left] {#5} ($(h2)!#1!-90:(h1)$); 
    \draw[#6]($(h1)!#1!-90:(h2)$)-- node [auto=right] {#7} ($(h2)!#1!90:(h1)$);
    }

    \begin{tikzpicture}[myn/.style={circle,  thin, draw, inner sep=0.15cm, outer sep=2pt,  fill=my_cp_col5}]


    \node[myn] (s) at (0,2) {\textcolor{my_cp_col1}{$1$}};
    \path (s) + (0,0.8) coordinate (c);
    \node (x1) at (c)  {$\mathbf{x}(1)$};

    \node[myn] (a) at (2,3.5) {\textcolor{my_cp_col1}{$2$}};
    \path (a) + (0,0.8) coordinate (c);
    \node (x2) at (c)  {$\mathbf{x}(2)$};

    \node[myn] (b) at (2,0.5) {\textcolor{my_cp_col1}{$3$}};
    \path (b) + (0,-0.8) coordinate (cx);
    \node (x3) at (cx)  {$\mathbf{x}(3)$};

    \node[myn] (c) at (5,3.5) {\textcolor{my_cp_col1}{$4$}};
     \path (c) + (0,0.8) coordinate (cx);
     \node (x4) at (cx)  {$\mathbf{x}(4)$};

    \node[myn] (d) at (5,0.5) {\textcolor{my_cp_col1}{$5$}};
     \path (d) + (0,-0.8) coordinate (cx);
     \node (x5) at (cx)  {$\mathbf{x}(5)$};

    \node[myn] (t) at (7,2) {\textcolor{my_cp_col1}{$6$}};
    \path (t) + (0,0.8) coordinate (cx);
    \node (x6) at (cx)  {$\mathbf{x}(6)$};


\path (a) + (-1.5,0.8) coordinate (caux);
\node[color=my_cp_col4] at (caux)  { $\mathbf{S}_{1}\rightarrow \mathcal{E}_1$ };

\path (d) + (1.5,-0.8) coordinate (caux);
\node[color=my_cp_col2] at (caux)  { $\mathbf{S}_{2}\rightarrow \mathcal{E}_2$ };


    \DoubleLine{s}{a}{-,very thick, my_cp_col4}{1}{-,very thick, my_cp_col2}{7}
    \DoubleLine{s}{b}{-,very thick, my_cp_col4}{2}{-,very thick, my_cp_col2}{6}
    \DoubleLine{a}{b}{-,very thick, my_cp_col4}{3}{-,very thick, my_cp_col2}{5}
    \DoubleLine{a}{c}{-,very thick, my_cp_col4}{4}{-,very thick, my_cp_col2}{4}
    \DoubleLine{b}{d}{-,very thick, my_cp_col4}{5}{-,very thick, my_cp_col2}{3}
    \DoubleLine{c}{t}{-,very thick, my_cp_col4}{6}{-,very thick, my_cp_col2}{2}
    \DoubleLine{d}{t}{-,very thick, my_cp_col4}{7}{-,very thick, my_cp_col2}{1}
    \DoubleLine{c}{d}{-,very thick, my_cp_col4}{7}{-,very thick, my_cp_col2}{1}
    

    \end{tikzpicture}

%% file: v25/sec_Spectral_Representation_Filters.tex

\section{Spectral Representations of Non Commutative Filters}\label{sec:spectopt}

In this section we discuss spectral representations of filters and signals. We will show that spectral representations in non commutative signal models are determined by matrix polynomials and that frequencies themselves are described by matrices instead of scalars. This introduces technical challenges, and yet we will also show that despite the significant differences with respect to the commutative scenario, spectral filtering is an analog of spectral filtering in commutative signal models. Additionally, this analog behavior extends to the norms of the operators, in particular the norm of non commutative filters is determined by the norm of the spectral responses which can be characterized in analogous ways to those in commutative signal models when considering induced norm on direct sums of spaces.


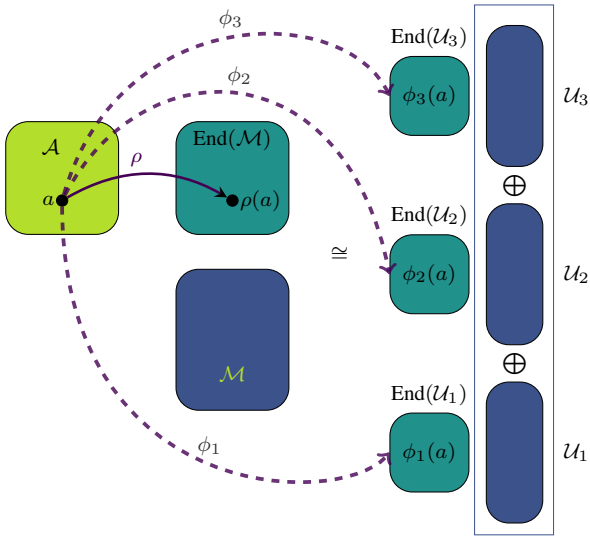
\begin{figure}
	\centering \input{./figures/fig_1_tikz_source.tex}
	\caption{Spectral decomposition in an ASM: The maps $\phi_i$ indicate the frequencies while each $\ccalU_i$ is an eigenspace. Notice that if $a\in\ccalA$ is a polynomial in terms of the generators of $\ccalA$, then $\rho(a)$ and $\phi_i (a)$ are polynomial operators in terms of the shift operators in $\End{M}$ and $\text{End}(\ccalU_i)$, respectively. When $\dim (\ccalU_i)=1$, $\phi_i (a)$ are polynomials in terms of the $i$th eigenvalue of the operator $\rho(a)$.}
	\label{fig_3}
\end{figure}


The notion of spectral decomposition descends from the concepts of irreducible and completely reducible subrepresentations of an algebra~\cite{algSP0, folland2016course, repthybigbook,terrasFG,diaconis1988group}. Widely
known notions of spectral decompositions used in graph signal
processing (GSP), discrete signal processing (DSP), and discrete
time signal processing (DTSP) among others, are obtained as
particular cases of a decomposition of representations of an
algebra as a sum of irreducible subrepresentations. Then, in order
to present a natural generalization of frequency decompositions
in non commutative ASM, we discuss the concepts of
subrepresentation, irreducibility and decomposability. We restrict
our attention to algebras that have a finite number of generators.


\begin{definition}
	
Let $(\mathcal{M},\rho)$ be a representation of $\mathcal{A}$. Then, a representation $(\mathcal{U},\rho)$ of $\mathcal{A}$ is a subrepresentation of  $(\mathcal{M},\rho)$ if $\mathcal{U}\subseteq\mathcal{M}$ and $\mathcal{U}$ is invariant under all operators $\rho(a)~\forall~a\in\mathcal{A}$, i.e. $\rho(a)u\in\mathcal{U}$ for all $u\in\mathcal{U}$ and $a\in\mathcal{A}$. 

\end{definition}


The notion of subrepresentation is tied to the property of invariance since subrepresentations are invariant under the action of the instantiations of elements of $\ccalA$ in $\End{M}$. Notice that when considering representations of an algebra with a single generator, the spaces generated by subsets of eigenvectors  of $\rho(a)$ determine the subrepresentations of $(\ccalM,\rho)$.

Now, we introduce formally irreducible subrepresentations which provide a minimal structural unit of invariance.


\begin{definition}
	
A representation $(\mathcal{M},\rho)$ (with $\ccalM\neq 0$) is irreducible or simple if the only subrepresentations of $(\mathcal{M},\rho)$ are $(0,\rho)$ and $(\mathcal{M},\rho)$.
	
\end{definition}


The irreducibility property of a subrepresentation implies that  there is not a subspace that has its own invariance under the action of the elements of the algebra. Additionally, notice that one dimensional subrepresentations are always irreducible. It is important to point out that the calculation of irreducible subrepresentations entails a significant computational cost -- see Appendix~\ref{sec_spect_rep_multiplicty} in the supplementary material.

As we will show in the next subsection, irreducibility can be used to write a general representation as a \textit{decomposition} in terms of irreducible subrepresentations. To show this we introduce the notion of direct sum of representations. Given two representations $(\ccalM_1, \rho_1)$ and $(\ccalM_1, \rho_2)$ of an algebra $\ccalA$ we can obtain a new representation, called the \textit{direct sum representation}, as $(\ccalM_1 \oplus \ccalM_1, \rho)$ where $\rho(a)(u_1 \oplus u_2) = \rho_1 (a)u_1 \oplus \rho_2 (a)u_2 $.


\subsection{Fourier Decompositions and Spectral Representation of Filters}

Using the ideas and concepts discussed above we introduce the notion of Fourier decomposition in algebraic signal processing.


\begin{definition}[Fourier Decomposition]\label{def:foudecomp}
For an algebraic signal model $(\mathcal{A},\mathcal{M},\rho)$ we say that there is a spectral or Fourier decomposition if
\begin{equation}
(\mathcal{M},\rho)\cong\bigoplus_{(\mathcal{U}_{i},\phi_{i})\in\text{Irr}\{\mathcal{A}\}}(\mathcal{U}_{i},\phi_{i})
,
 \label{eq:foudecomp1}
\end{equation}
where the $(\mathcal{U}_{i},\phi_{i})$ are irreducible subrepresentations of $(\mathcal{M},\rho)$. Any signal $\mathbf{x}\in\mathcal{M}$ can be therefore represented by the map $\Delta$ given by
\begin{equation}
\Delta: \mathcal{M} \to \bigoplus_{(\mathcal{U}_{i},\phi_{i})\in\text{Irr}\{\mathcal{A}\}}\mathcal{U}_{i}
\label{eq:foudecomp2}
\end{equation}
\begin{equation*}
\mathbf{x}\mapsto \hat{\mathbf{x}}
,
\end{equation*}
known as the Fourier decomposition of $\mathbf{x}$ and the projection of $\hat{\mathbf{x}}$ in each $\mathcal{U}_{i}$ are the Fourier components represented by $\hat{\mathbf{x}}(i)$.
\end{definition}


In Definition~\ref{def:foudecomp} it is assumed that each individual subrepresentation $(\ccalU_i , \phi_i)$ cannot be expressed as a direct sum of irreducible subrepresentations isomorphic to $(\ccalU_i , \phi_i)$. This assumption is analogous to the assumption of not having repeated eigenvalues when considering spectral decompositions where the irreducible subrepresentations have dimension equal to 1. This choice is done for the sake of simplicity and we refer the reader to Appendix~\ref{sec_spect_rep_multiplicty} of the supplementary material for a more sophisticated formulation of the Fourier decompositions. We remark that the decomposition in~\eqref{eq:foudecomp1} and~\ref{eq:foudecomp2} is unique up to isomorphism -- see Appendix~\ref{sec_spect_rep_multiplicty} in the supplementary material.

It is worth pointing out that the homomorphisms $\phi_i$ associated to each non isomorphic irreducible representation define the \textit{frequency} associated to the vector space $\ccalU_i$~\cite{algSP0} -- see Fig.~\ref{fig_3}. In commutative scenarios, like for instance in GSP, we have $\dim (\ccalU_i) = 1$. In this case the term $\phi_i (a)$ is a scalar value which corresponds to the eigenvalues of $\rho(a)$ while $\ccalU_i$ is the space spanned by the $i$-th eigenvector.\footnote{We refer the reader to Appendix G in the supplementary material where we show with more details how the Fourier transform in GSP can be obtained as a particular case of Definition~\ref{def:foudecomp}.}. If $\dim (\ccalU_i) > 1$ then $\phi_i (a)$ is a matrix. This last scenario is typical when non commutative algebras are involved.


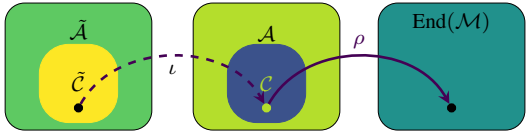
\begin{figure}
	\centering
	\input{./figures/50_asp_FilterClasses_2.tex}	
	\caption{Schematic representation of the spectral representation $\tilde{\ccalC}$ of a subset of filters $\ccalC\subset\ccalA$
		 algebraic signal model. The properties of $\rho(\ccalC)$ are determined by $\ccalC$ which is characterized by the elements in $\tilde{\ccalC}$.}
	\label{fig_filtpropdom}
\end{figure}


We synthesize the role of Fourier decompositions and the filtering operation in the following theorem.


\begin{theorem}(Filtering Spectral Theorem)\label{thm_filtspec}
Let $(\ccalA, \ccalM, \rho)$ be an algebraic signal model where $\ccalA$ has generators $\{ g_k \}_{k=1}^{m}$, and $\rho(a)$ is the realization of $a\in\ccalA$ in $\End{M}$ by means of $\rho$. If $
(\mathcal{M},\rho)\cong\bigoplus_{(\mathcal{U}_{i},\phi_{i})\in\text{Irr}\{\mathcal{A}\}}(\mathcal{U}_{i},\phi_{i})$ then
\begin{multline}
\rho \left( 
                 p(g_1,\ldots, g_m)
        \right)
        =
       p \left( 
                  \bbS_1, \ldots, \bbS_m
          \right)  
       =       
       \\
       \sum_{i}
                 p \left( 
                 \boldsymbol{\Lambda}_{1,i},
                 \ldots,
                 \boldsymbol{\Lambda}_{m,i}
                 \right)
                 \mathrsfso{P}_{i}
                 ,
\end{multline}
where $\mathrsfso{P}_{i}$ is the projection operator on $\ccalU_i$ and $ \boldsymbol{\Lambda}_{k,i} = \phi_{i}(g_k)$. Additionally, if $\bby=\rho\left( p(g_1, \ldots, g_m)  \right)\bbx$, then
\begin{equation}
\hat{\bby}_i 
              =
              p \left( 
                          \boldsymbol{\Lambda}_{1,i},
                          \ldots,
                          \boldsymbol{\Lambda}_{m,i}
                 \right)
              \hat{\bbx}_i   
              ,
\end{equation}
where the sub index $i$ indicates a projection on $\ccalU_i$.

\end{theorem}


\begin{proof}
	See Appendix~\ref{proof_thm_spec_filt} in the supplementary material.
\end{proof}

Theorem~\ref{thm_filtspec} exhibits similarties in form with the classical spectral theorem for commutative signal models but also substantial differences. Indeed, Theorem~\ref{thm_filtspec} is a generalization of the classical spectral theorem where decompositions of operators are expressed in terms of projections on spaces spanned by more than just one vector. As indicated in Appendix~\ref{proof_thm_spec_filt}, Theorem~\ref{thm_filtspec} follows from the fact that the restriction of the homomorphism $\rho$ to an invariant subspace $\ccalU_i$ given by $\phi_i$ is again a homomorphism. Therefore, if $p(g_1, \ldots, g_m)$ is a polynomial so it is $\rho(p(g_1, \ldots, g_m))$ and $\phi_i(p(g_1, \ldots, g_m))$ -- and they have the same coefficients. Additionally, the dimensions of $\ccalM$ and $\ccalU_i$ will determine the dimensions of the independent variables in $\rho(p(g_1, \ldots, g_m))$ and $\phi_i(p(g_1, \ldots, g_m))$.

The algebraic filter $a\in\ccalA$ determines the form and properties of the spectral response indicated by the homomorphisms $\phi_i$. Therefore, classes of filters in $\ccalA$ lead to specific classes of filters defined in the spectral domain. We use this fact and an auxiliary matrix algebra, $\tilde{\ccalA}$, to provide a concrete characterization of subsets of filters in $\ccalA$. $\tilde{\ccalA}$ is endowed with a norm and is isomorphic to $\ccalA$ -- see Fig.~\ref{fig_filtpropdom}.

In $\tilde{\ccalA}$ the generators can be considered as variables taking specific values on a vector space. In particular, given any $(\ccalA, \ccalM, \rho)$ with $\ccalA$ generated by $\{ g_{i}\}_{i=1}^{m}$ we can associate 
 $
 \tilde{\ccalA} = 
 \left\lbrace \left.
 p: M_{r,r}^{m}(\mathbb{C}) \mapsto M_{r,r}(\mathbb{C})
 \right\vert
 p: \text{polynomial}
 \right\rbrace,        
 $
where $M_{r,r}(\mathbb{C})$ is the space of matrices of size $r\times r$ whose entries belong to $\mathbb{C}$, while $M_{r,r}^{m}(\mathbb{C})$ is the $m$-times cartesian product of $M_{r,r}(\mathbb{C})$. Additionally, we endow $\tilde{\ccalA}$ with a norm. It is possible to see that there is a natural isomorphism $\iota$ between $\ccalA$ and $\tilde{\ccalA}$ given by $\iota (g_{i}) = \boldsymbol{\Lambda}_{i}\in M_{r,r}(\mathbb{C})$. We attribute the properties of a set of filters realized in the set $\tilde{\ccalC}\subset\tilde{\ccalA}$ to the set
%
%
%
$
\ccalC
=
\left\lbrace
a\in\ccalA
\vert
\quad\iota(a)\in\tilde{\ccalC}                                      
\right\rbrace.
$
%
%
%
The value of $r$ is selected as 
%
%
%
$
r  
  =
  \max
       \left\lbrace 
             d_i
       \right\rbrace
       ,
$
%
%
%
where $d_i$ are the dimensions of the irreducible subrepresentations of $(\ccalM,\rho)$. It is worth pointing out that the value of the $d_i$ depends on $\rho$, and in general $d_i \ll \dim (\ccalM)$~\cite{repthysmbook,repthybigbook,parada_algnn,algSP0}. Taking into account this, we formally introduce the spectral representation of a filter.


\begin{definition}\label{definition:filtspecrep}
	Let $(\ccalA, \ccalM, \rho)$ be an ASM where $\ccalA$ is an algebra with generators $\{ g_{i}\}_{i=1}^{m}$. Let $\tilde{\ccalA}$ the matrix algebra isomorphic to $\ccalA$ given by
	\begin{equation}
	\tilde{\ccalA} = 
	\left\lbrace \left.
	p(\boldsymbol{\Lambda}_1,\ldots,\boldsymbol{\Lambda}_m): M_{r,r}^{m}(\mathbb{C}) \mapsto M_{r,r}(\mathbb{C})
	\right\vert
	p: \text{polynomial}
	\right\rbrace,                       
	\end{equation}
	where $r  
	=
	\max
	\left\lbrace 
	d_i
	\right\rbrace
	,$ and $d_i$ are the dimensions of the irreducible subrepresentations of $(\ccalM,\rho)$.
	Then, we say that $p(\boldsymbol{\Lambda}_1,\ldots,\boldsymbol{\Lambda}_m)\in\tilde{\ccalA}$ is the spectral representation of $p(g_1,\ldots,g_m)\in\ccalA$.
\end{definition}



\begin{remark}\normalfont
 It is important to remark that a substantial part of the technical challenges that have made the analysis of non commutative signal models and architectures elusive, is precisely the difficulty associated to the use of the Fourier representations. As we just discussed, some frequencies are not described by scalars but instead by matrices. Part of our contribution in this paper is precisely the fact that we have overcome these technical challenges and as the reader can corroborate in Appendix~\ref{sec:proofofTheorems} the formal proof of our stability results make use of tools different from those used in~\cite{parada_algnn} for commutative signal models where frequencies are associated to scalars. 
\end{remark}


As a final comment it is worth highlighting that the notion of aliasing on the spectral domain for non commutative signal models is analog to that one of commutative models. The difference is that the representation of the aliased information is expressed in terms the basis of the $\ccalU_i$ spaces where aliasing takes place.

%% file: figures/fig_1_tikz_source.tex

\def \scale { 1.5}
\def \unit  { \scale cm}


\definecolor{my_cp_col1}{RGB}{253, 231, 37}
\definecolor{my_cp_col2}{RGB}{180, 222,44}
\definecolor{my_cp_col3}{RGB}{94, 201, 98}
\definecolor{my_cp_col4}{RGB}{33, 145, 140}
\definecolor{my_cp_col5}{RGB}{59, 82, 139}
\definecolor{my_cp_col6}{RGB}{68, 1, 84}


\tikzstyle{set} = [ rectangle,
rounded corners = 0.2*\unit,
inner sep=0pt,
draw,
anchor = center ]

\tikzstyle{vector space} = [ set,
fill=my_cp_col5,
minimum width  = 1*\unit,
minimum height = 1.25*\unit]

\tikzstyle{vectorspace2} = [ set,
fill=my_cp_col5,
minimum width  = 0.5*\unit,
minimum height = 1.25*\unit]

\tikzstyle{endomorphisms} = [ vector space,
fill=black!15,
minimum height = 1*\unit,
minimum width = 1*\unit]

\tikzstyle{endomorphisms2} = [vector space,
fill=my_cp_col4,
minimum height = 0.7*\unit,
minimum width = 0.7*\unit]

\tikzstyle{algebra} = [ endomorphisms,
fill=my_cp_col2,
minimum width = 1*\unit,
minimum height = 1*\unit]

\tikzstyle{dot} = [ circle,
minimum width  = 0.1*\unit,
fill=black,
inner sep=0pt,
draw,
anchor = center ]

{\fontsize{8}{8}\selectfont
	
	\begin{tikzpicture}[-stealth, draw = black!99, scale = \scale]

	
	\path (0,0) node [vector space,opacity=1] (M) {};
	\path (M.south) ++ (0, 0.2) node [above] {\textcolor{my_cp_col2}{$\ccalM$}};

   
   \path (M.south)+(2.5,-1) node [vectorspace2,opacity=1,anchor=south] (V1) {};
   \path (V1.east) ++ (0.3, 0) node []  {$\ccalU_1$};
   \path (V1.north) node [opacity=1,anchor=south] (bigsum1) {$\bigoplus$};
    
   \path (bigsum1.north)+(0,0) node [vectorspace2,opacity=1,anchor=south] (V2) {};
   \path (V2.east) ++ (0.3, 0) node []  {$\ccalU_2$};
   \path (V2.north) node [opacity=1,anchor=south] (bigsum2) {$\bigoplus$};
   
   \path (bigsum2.north)+(0,0) node [vectorspace2,opacity=1,anchor=south] (V3) {};
   \path (V3.east) ++ (0.3, 0) node []  {$\ccalU_3$};     

   \path (V1.south west) + (-0.1,-0.1) coordinate (caux);
   \draw[color=my_cp_col5,line width=0.1mm] (caux) rectangle +(0.7,4.7);

   
   \path (V1.west) ++ (-0.5, 0) node [endomorphisms2] (EndV1) {$\phi_1 (a)$};
   \path (EndV1.north) ++ (0, 0) node [anchor=south]  {$\text{End}(\ccalU_1)$};
   
   \path (V2.west) ++ (-0.5, 0) node [endomorphisms2] (EndV2) {$\phi_2 (a)$};
   \path (EndV2.north) ++ (0, 0) node [anchor=south]  {$\text{End}(\ccalU_2)$};
   
   \path (V3.west) ++ (-0.5, 0) node [endomorphisms2] (EndV3) {$\phi_3 (a)$};
   \path (EndV3.north) ++ (0, 0) node [anchor=south]  {$\text{End}(\ccalU_3)$};

	
	\path (M.north) ++ (0, 0.3) 
	node [endomorphisms, anchor=south,fill=my_cp_col4] (End) {};
	\path (End.north) ++ (0.0, 0) node [below] {$\text{End}(\ccalM)$};   
	
	\path (End.center) ++ (0,-0.2) node [dot] (e) {};      
	\path (e) node [right] {$\rho(a)$};

    
    \path (M.north east) ++ (0.3, 0) node [anchor=south west]  {$\cong$};

	
	\path (End.north west) ++ (-0.5, 0) 
	node [algebra, anchor = north east,opacity=1] (A) {};   
	\path (A.north) ++ (-0.1,-0.1) node [below] {$\ccalA$};   
    
	\path (A.center) ++ (0,-0.2) node [dot] (a) {};      
	\path (a) node [left] {$a$};


\draw [->,line width=0.5mm, opacity=0.8,color=my_cp_col6,dashed] (a)..controls (-1,3) and (1,3)..(EndV3.west) node[midway,above,rotate=0,color=black] {$\phi_{3}$};
	
\draw [->,line width=0.5mm, opacity=0.8,color=my_cp_col6,dashed] (a)..controls (-0.8,2.6) and (1,2.6)..(EndV2.west) node[midway,above,rotate=0,color=black] {$\phi_{2}$};	
	
\draw [->,line width=0.5mm, opacity=0.8,color=my_cp_col6,dashed] (a)..controls (-1.5,-1.5) and (1,-1.5)..(EndV1.west) node[midway,above,rotate=0,color=black] {$\phi_{1}$};

   \path (End.west) + (-0.1,0.25) coordinate (c1);
   \path (End.east) + (+0.1,0.25) coordinate (c2);   
   \path [draw, -stealth, line width = 1.0, my_cp_col6] 
         (a) edge [bend left] node [above] {$\rho~~~$} (e);

	\end{tikzpicture}
	
}

%% file: figures/50_asp_FilterClasses_2.tex


\definecolor{my_alejocol5a}{RGB}{1,31,75}
\definecolor{my_alejocol4a}{RGB}{3,57,108}
\definecolor{my_alejocol3a}{RGB}{0,91,150}
\definecolor{my_alejocol2a}{RGB}{100,151,177}
\definecolor{my_alejocol1a}{RGB}{179,205,224}

\definecolor{my_alejocol4b}{RGB}{212, 220, 220}
\definecolor{my_alejocol3b}{RGB}{113, 18, 55}
\definecolor{my_alejocol2b}{RGB}{236, 85, 141}
\definecolor{my_alejocol1b}{RGB}{225, 73, 132}

\definecolor{my_alejocol4c}{RGB}{81, 185, 239}
\definecolor{my_alejocol3c}{RGB}{119, 164, 211}
\definecolor{my_alejocol2c}{RGB}{193, 223, 213}
\definecolor{my_alejocol1c}{RGB}{53, 89, 85}

\colorlet{my_alejocolg1}{black!30}
\colorlet{my_alejocolg2}{black!35}
\colorlet{my_alejocolg3}{black!40}
\colorlet{my_alejocolg4}{black!45}
\colorlet{my_alejocolg5}{black!50}
\colorlet{my_alejocolg6}{black!55}
\colorlet{my_alejocolg7}{black!60}

\definecolor{my_cp4_col1}{RGB}{255, 86, 87}
\definecolor{my_cp4_col2}{RGB}{55, 108, 138}
\definecolor{my_cp4_col3}{RGB}{242, 217, 187}
\definecolor{my_cp4_col4}{RGB}{99, 143, 169}

\definecolor{my_cp5_col1}{RGB}{7, 117, 232}
\definecolor{my_cp5_col2}{RGB}{144, 171, 229}
\definecolor{my_cp5_col3}{RGB}{30, 73, 164}
\definecolor{my_cp5_col4}{RGB}{158, 158, 100}
\definecolor{my_cp5_col5}{RGB}{193, 195, 199}
\definecolor{my_cp5_col6}{RGB}{83, 83, 83}

\definecolor{my_cp_col1}{RGB}{253, 231, 37}
\definecolor{my_cp_col2}{RGB}{180, 222,44}
\definecolor{my_cp_col3}{RGB}{94, 201, 98}
\definecolor{my_cp_col4}{RGB}{33, 145, 140}
\definecolor{my_cp_col5}{RGB}{59, 82, 139}
\definecolor{my_cp_col6}{RGB}{68, 1, 84}

\usetikzlibrary{positioning,decorations.pathreplacing,shapes}


\def \scale {1.3}
\def \unit  { \scale cm}
\def \layerinterdist {2.5}

\tikzstyle{set} = [rectangle,color=black,
                    rounded corners = 0.2*\unit,
                    fill=black,
                    inner sep=0pt,
                    draw,
                    anchor = center,
                    line width=0.1mm]

\tikzstyle{vectorspace} = [ set, 
                             fill=my_alejocol4c!50,
                             minimum width  = 1.5*\unit,
                             minimum height = 2.0*\unit]
                             
\tikzstyle{endomorphisms} = [ vectorspace,
                              fill=black!15,
                              minimum height = 1.3*\unit]
                                
\tikzstyle{subendomor2in1} = [endomorphisms, rounded corners=0.3*\unit,
                              fill=my_cp_col1,
                              minimum width  = 0.8*\unit,
                              minimum height = 0.8*\unit
                             ]                                                                                                                                                
\tikzstyle{algebra} = [ endomorphisms,
                        fill=my_alejocol2b!50,
                        minimum width = 1.5*\unit]

\tikzstyle{dot} = [ circle,
                    minimum width  = 0.05*\unit,
                    fill=black,
                    color=black,
                    inner sep=0pt,
                    draw,
                    anchor = center ]

{\fontsize{8}{8}\selectfont

\begin{tikzpicture}[rounded corners,ultra thick]


   \path (0,0) node [] (M0) {};

   \path (M0.north) ++ (0, 0.5) 
         node [endomorphisms, anchor=south, fill=my_cp_col4] (End0) {};
   \path (End0.north) ++ (0.0, 0) node [below, color=black] {$\text{End}(\mathcal{M})$};

   \path (End0.south) ++ (0.0, 0.3) node [dot] (e) {};      

   \path (End0.north west) ++ (-3, 0) 
         node [algebra, anchor = north east, fill=my_cp_col3] (At) {};   
   \path (At.north) ++ (0,-0.1) node [below,color=black] {$\tilde{\mathcal{A}}$};  
   
    \path (At.south) ++ (0,0.1) 
    node [subendomor2in1, anchor=south,color=my_cp_col1] (subset2) {};
    \path (subset2.south) ++ (0, 0.3) node [above, color=black] {$\tilde{\mathcal{C}}$};
    
    \path (subset2.south) ++ (0.0, 0.2) node [dot] (a2) {}; 
   
   \path (End0.north west) ++ (-0.5, 0) 
         node [algebra, anchor = north east, fill=my_cp_col2] (A) {};   
   \path (A.north) ++ (0,-0.2) node [below,color=black] {$\mathcal{A}$};  
    
    \path (A.south) ++ (0,0.1) 
    node [subendomor2in1, anchor=south,color=my_cp_col5] (subset2) {};
    \path (subset2.south) ++ (0, 0.3) node [above, color=my_cp_col2] {$\mathcal{C}$};
    
    \path (subset2.south) ++ (0.0, 0.2) node [dot,color=my_cp_col2] (a1) {}; 
        
   \path (A.north) + (0.5,-0.5) coordinate (c1);
   \path (End0.north) + (-0.5,-0.5) coordinate (c2);   
   \path [draw, -stealth,line width = 1.0,color=my_cp_col6] (a1) .. controls (c1) and (c2) .. (e) node[midway,above ,rotate=0,color=my_cp_col6] {$\rho$};
   
   \path (At.north) + (0.5,-0.5) coordinate (c1); 
   \path (A.north) + (-0.5,-0.5) coordinate (c2);
   \path [->,draw, -stealth,line width = 1.0,color=my_cp_col6,dashed] (a2) .. controls (c1) and (c2) .. (a1) node[midway,below,rotate=0,color=black] {$\iota$};

\end{tikzpicture}

}

%% file: v25/sec_algNN.tex

%
%

\section{Algebraic Neural Networks with Non Commutative Algebras}\label{sec_Algebraic_NNs}


\begin{figure}
	\centering\input{./figures/gnn_block_diagram_single_feature_uncolored_01.tex}
	\caption{Algebraic Neural Network $\{ (\ccalA_{\ell},\ccalM_{\ell},\rho_{\ell}; \sigma_\ell ) \}_{\ell=1}^{3}$ with three layers indicating how the input signal $\mathbf{x}$ is processed  and mapped into $\mathbf{x}_{3}$. In each layer the information is transformed by a convolutional filter $\rho_{\ell}(a_\ell)$, followed by a pointwise nonlinearity $\eta_\ell$ and a pooling operator $P_{\ell}$.}
	\label{fig_6}
\end{figure}
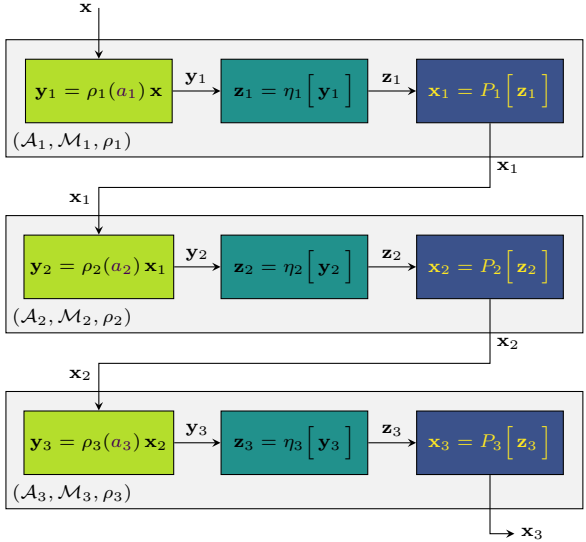


Algebraic neural networks (AlgNNs) are stacked layered structures (see Fig.~\ref{fig_6}) where the processing of information in each layer is carried out by means of filters of an algebraic signal model and pointwise nonlinear operators. For a training set $\ccalT = \{ \bbx, \bby \}$ with inputs $\bbx$ and outputs $\bby$, it is possible to learn the algebraic filters $a_\ell \in\ccalA_\ell$ in each layer of the AlgNN to
produce a mapping representation. This allows an estimation of
the output to an unseen input $\tilde{\bbx}\in\ccalT$. The data from the training set $\ccalT$ is used to find subsets of filters $\ccalP_{\ell}\subset \ccalA_{\ell}$
that minimize a cost function of the form 
$
\sum_{(\bbx,\bby)\in\ccalT} f_{\ccalP_\ell}(\bbx,\bby),
$
where $f_{\ccalP_\ell}(\bbx,\bby)$ is a fitting metric that penalizes the difference
between $\bby$ and the output of the AlgNN produced when the input
is $\bbx$. In the $\ell$-th layer of the AlgNN, an incoming signal $\bbx_{\ell-1}$ from the
layer $\ell-1$ is filtered by means of the convolution $\rho_{\ell}(a_\ell)\bbx_{\ell-1} $. Then, a pointwise nonlinear operator $\eta_{\ell}: \ccalM_{\ell}\rightarrow\ccalM_{\ell }$ is applied and finally a pooling operator $P_{\ell}: \ccalM_{\ell} \rightarrow \ccalM_{\ell+1}$ matches information between $\ccalM_\ell$ and $\ccalM_{\ell+1}$. The
output signal of the layer $\ell$ can be written as
\begin{equation}\label{eq:xl}
\bbx_{\ell}=\sigma_{\ell}\left(\rho_{\ell}(a_{\ell})\bbx_{\ell-1}\right)=\Phi (\mathbf{x}_{\ell-1},\mathcal{P}_{\ell},\mathcal{S}_{\ell})
.
\end{equation}
We use the symbol $\Phi (\mathbf{x}_{\ell-1},\mathcal{P}_{\ell},\mathcal{S}_{\ell})$ to make emphasis in the fact that the filters used in each layer belong to specific subsets of the algebra and
that a specific family of shift operators is being used. If several features per layer are used, we use the notation
\begin{equation}\label{eqn:alg_feat}
\bbx_{\ell}^{f}=\sigma_{\ell}\left(\sum_{g=1}^{F_{\ell}}\rho_{\ell}\left(a_{\ell}^{gf}\right)\bbx_{\ell-1}^{g}\right),
\end{equation}
where the super index $f$ indicates the $f$th feature and $\ell$ the layer were information is being processed. To denote an AlgNN with $L$ layers we use $\left\lbrace\left(\ccalA_{\ell},\ccalM_{\ell},\rho_{\ell}; \sigma_\ell \right)\right\rbrace_{1}^{L}$, where $\sigma_\ell =  P_\ell \eta_\ell$. As pointed out in~\cite{parada_algnn}, traditional neural networks (CNNs), graph neural networks (GNNs), and graphon neural networks (WNNs) among others, can be obtained as particular instantiations of a general AlgNN with a commutative algebra. For our discussion we consider that $\sigma_\ell = P_\ell \eta_\ell$ is $C_\ell$-Lipschitz and $\sigma_\ell (0) = 0$. The role of $\eta_\ell$ is crucial for the performance of any AlgNN, since it is the nonlinearity of $\eta_{\ell}$ what allows the AlgNN to redistribute spectral information and compensate restrictions imposed
on the filters. There are many possible choices for $\eta_\ell$ that could improve the generalization capacity of the
AlgNN. Typically a low computational cost $\eta_\ell$ is selected. The operator $P_{\ell}: \ccalM_{\ell}\rightarrow\ccalM_{\ell+1}$ performs an operation of
dimensionality reduction. In particular instantiations of AlgNNs
like GNNs, such operator can be associated to graph coarsening
techniques or optimal sampling strategies for bandlimited signals~\cite{gamagnns}. In what follows we present some examples of concrete non commutative convolutional architectures instantiated as particular cases of non commutative algebraic neural networks (AlgNNs).


\begin{remark}\normalfont

Note that the action of the pointwise nonlinearity, $\eta_\ell$, and the pooling operator, $P_\ell$, is defined in terms of a basis of $\ccalM_\ell$. This is, given a basis $\displaystyle\left\lbrace b_{i}^{(\ell)} \right\rbrace_i$ of $\ccalM_\ell$ we have 
$\eta_{\ell} \left( \sum_{i}\alpha_i b_{i}^{(\ell)} \right) = \sum_{i}\eta_{\ell} (\alpha_i)b_{i}^{(\ell)}$, and $P_{\ell}: \text{span}(\{ b_{i}^{(\ell)} \}_i) \to \text{span}(\{ b_{i}^{(\ell+1)} \}_i)$.
In our analysis, the specifics of such basis are not relevant as long as the operator $\sigma_\ell = P_\ell \eta_\ell$ is Lipschitz and $\sigma_\ell (0) = 0$. Notice also that since $P_\ell$ performs dimensionality reduction, there are multiple choices for linear and nonlinear versions of $P_\ell$ depending on the properties of $\ccalM_\ell$~\cite{boureau_pooling_1,boureau_pooling_2,ParadaMayorga2022GraphonPF} and we consider those choices for which $\sigma_\ell = P_\ell \eta_\ell$ is Lipschitz and $\sigma_\ell (0) = 0$. We point out however, that the optimal selection of such basis and the operators $\eta_\ell$ and $P_\ell$ opens up an interesting future research direction.

\end{remark}



\begin{example}[Multigraph neural networks]\normalfont \label{ex_multigraph_nn}
		Let us consider a multigraph $G=(\ccalV, \{ \ccalE_r \}_{r=1}^{m})$ with set of nodes $\ccalV$, $\vert \ccalV\vert =N$ and multiset of edges $\{ \ccalE_r \}_{r=1}^{m}$. Let $\bbS_i$ be a matrix representation of the multigraph on the set of edges $\ccalE_i$, which could be the adjacency matrix or a Laplacian matrix of the graph $( \ccalV, \ccalE_i )$. Then, the $\ell$-th layer of a multigraph neural network is composed by convolutional operators followed by a pointwise nonlinearity and a pooling operator. The convolutional filters are given by the multivariable polynomials (see Example~\ref{ex_multgsp}):
		$
		p(\bbS_1, \ldots, \bbS_m).
		$
		Then, given the input $\bbx_\ell$ to the $\ell$-th layer of the multigraph neural network, we leverage symmetries on the multigraph to obtain $\bby_\ell = p(\bbS_1, \ldots, \bbS_m)\bbx_\ell$. After this, we apply a pointwise nonlinear operator $\eta_\ell: \mbR^N \rightarrow \mbR^N$ to obtain $\bbz_\ell = \eta_\ell \left( \bby_\ell \right)$, with $\bbz_\ell (u) = \max\left\lbrace \bby_\ell (u), 0\right\rbrace$, and where $\bbz_\ell (u)$ is the $u$-th component of $\bbz_\ell$. Besides the convolutional filter and the pointwise nonlinear operator, an operation of pooling may be considered to reduce the computational cost of processing with multiple filters. For multigraph neural networks, the pooling operator may be defined using the zeroing approach used for GNNs in~\cite{gamagnns}, where the information is forced to be zero in a subset of nodes. This is done under the hypothesis that the information in those nodes is less relevant or redundant. Then, if zeroing is used the pooling operator is given by $P_\ell: \mbR^N \rightarrow \mbR^N$ with $\bbx_{\ell+1} = P_{\ell}( \bbz_\ell  )$ and $\bbx_{\ell+1} (u) =0$ for any $u\in\ccalU\subset\ccalV$, and where the subset $\ccalU$ could be chosen according to a specific heuristic or the minimization of a cost function. Notice that for the sake of simplicity, in this example we considered one filter per layer, but we can indeed use several filters.
\end{example}	


We refer the reader to Appendix~\ref{sec_grsp_discussion} and Appendix~\ref{sec_quaternions_sp_nn} for a discussion on how convolutional group neural networks and quaternion neural networks can be seen as particular cases of a generic AlgNN.


\begin{remark}\normalfont
    
The cost of the convolution in the $\ell$-th layer is $\mathcal{O}(N_{\ell}^2 m^K F_\ell G_{\ell})$, where $N_\ell = \text{dim}(\ccalM_\ell)$ is the dimension of the vector space, $m$ is the number of generators in the algebra, $F_{\ell}$ is the number of input features, $G_\ell$ is the number of output features, and $K$ is the order of the polynomial filters. The number of learnable parameters in each layer is $\mathcal{O}(m^K F_\ell G_{\ell})$, which does not depend on the dimension of the vector space. It is important to remark that the complexity associated with the learnable parameters can be reduced when adapting pruning algorithms like the one proposed in~\cite{Butler2022ConvolutionalLO,msp_icassp2023}, which reduces efficiently the number of monomials in an polynomial operator.

\end{remark}


%% file: figures/gnn_block_diagram_single_feature_uncolored_01.tex


\definecolor{my_alejocol5a}{RGB}{1,31,75}
\definecolor{my_alejocol4a}{RGB}{3,57,108}
\definecolor{my_alejocol3a}{RGB}{0,91,150}
\definecolor{my_alejocol2a}{RGB}{100,151,177}
\definecolor{my_alejocol1a}{RGB}{179,205,224}

\definecolor{my_alejocol4b}{RGB}{212, 220, 220}
\definecolor{my_alejocol3b}{RGB}{113, 18, 55}
\definecolor{my_alejocol2b}{RGB}{236, 85, 141}
\definecolor{my_alejocol1b}{RGB}{225, 73, 132}

\definecolor{my_alejocol4c}{RGB}{81, 185, 239}
\definecolor{my_alejocol3c}{RGB}{119, 164, 211}
\definecolor{my_alejocol2c}{RGB}{193, 223, 213}
\definecolor{my_alejocol1c}{RGB}{53, 89, 85}

\colorlet{my_alejocolg1}{black!30}
\colorlet{my_alejocolg2}{black!35}
\colorlet{my_alejocolg3}{black!40}
\colorlet{my_alejocolg4}{black!45}
\colorlet{my_alejocolg5}{black!50}
\colorlet{my_alejocolg6}{black!55}
\colorlet{my_alejocolg7}{black!60}

\definecolor{my_cp4_col1}{RGB}{255, 86, 87}
\definecolor{my_cp4_col2}{RGB}{55, 108, 138}
\definecolor{my_cp4_col3}{RGB}{242, 217, 187}
\definecolor{my_cp4_col4}{RGB}{99, 143, 169}

\definecolor{my_cp5_col1}{RGB}{7, 117, 232}
\definecolor{my_cp5_col2}{RGB}{144, 171, 229}
\definecolor{my_cp5_col3}{RGB}{30, 73, 164}
\definecolor{my_cp5_col4}{RGB}{158, 158, 100}
\definecolor{my_cp5_col5}{RGB}{193, 195, 199}
\definecolor{my_cp5_col6}{RGB}{83, 83, 83}

\definecolor{my_cp_col1}{RGB}{253, 231, 37}
\definecolor{my_cp_col2}{RGB}{180, 222,44}
\definecolor{my_cp_col3}{RGB}{94, 201, 98}
\definecolor{my_cp_col4}{RGB}{33, 145, 140}
\definecolor{my_cp_col5}{RGB}{59, 82, 139}
\definecolor{my_cp_col6}{RGB}{68, 1, 84}

\def \myfactor {0.65}
\def \unit  {\myfactor cm}

\tikzstyle{block} = [ rectangle,
                      minimum width = \unit,
                      minimum height = \unit,
                      fill = gray,
                      draw = black,
                      text = black]

\tikzstyle{filter} = [block,
                      minimum width  = 3.0*\unit,
                      minimum height = 1.3*\unit,
                      fill=my_cp_col2]

\tikzstyle{nonlinearity} = [ filter,
                             minimum width  = 3.0*\unit,
                             fill = my_cp_col4]

\tikzstyle{pooling} = [ filter,
                             minimum width  = 3.0*\unit,
                             fill = my_cp_col5]

\def \deltainput     {( 0.0,-1.7)}
\def \deltaoutput    {( 0.0,-1.2)}
\def \deltalayer     {3.6}
\def \deltaconnector {1.45}
\def \deltasigma     {( 4, 0.0)}

\def \one   {$\displaystyle{\mathbf{y}_{1}  = \rho_{1}(\textcolor{my_cp_col6}{a_1})\,\mathbf{x}}$}
\def \two   {$\displaystyle{\mathbf{y}_2  =  \rho_{2}(\textcolor{my_cp_col6}{a_2})\,\mathbf{x}_{1}}$}
\def \three {$\displaystyle{\mathbf{y}_3  = \rho_{3}(\textcolor{my_cp_col6}{a_3})\,\mathbf{x}_{2}}$}
\def \sigmaone   {$\displaystyle{\mathbf{z}_{1} = {\eta_{1}} \Big[\, \mathbf{y}_1 \, \Big]}$}
\def \sigmatwo   {$\displaystyle{\mathbf{z}_{2} = {\eta_{2}} \Big[\, \mathbf{y}_2 \, \Big]}$}
\def \sigmathree {$\displaystyle{\mathbf{z}_{3} = {\eta_{3}} \Big[\, \mathbf{y}_3 \, \Big]}$}
\def \proyone{\textcolor{my_cp_col1}{$\displaystyle{\mathbf{x}_{1} = {P_{1}} \Big[\, \mathbf{z}_1 \, \Big]}$}}
\def \proytwo{\textcolor{my_cp_col1}{$\displaystyle{\mathbf{x}_{2} = {P_{2}} \Big[\, \mathbf{z}_2 \, \Big]}$}}
\def \proythree{\textcolor{my_cp_col1}{$\displaystyle{\mathbf{x}_{3} = {P_{3}} \Big[\, \mathbf{z}_3 \, \Big]}$}}

%
{\fontsize{7}{7}\selectfont\begin{tikzpicture}[scale = \myfactor]

  \pgfdeclarelayer{bg}     
  \pgfsetlayers{bg,main}   

  \node (input) [rectangle, minimum width = 0.1*\unit] {$\mathbf{x}$};
  \path (input.east)      ++ \deltainput node [filter]       (L1 Filter1) {\one};
  \path (L1 Filter1) ++ \deltasigma node [nonlinearity] (L1 F1)      {\sigmaone};
  \path (L1 F1) ++ \deltasigma node [pooling] (L1 F2)      {\textcolor{black}{\proyone}};
  \path[draw, -stealth] (L1 Filter1.east) -- node [above] {$\mathbf{y}_1$} (L1 F1.west);
  \path[draw, -stealth] (L1 F1.east) -- node [above] {$\mathbf{z}_1$} (L1 F2.west);

  \path (L1 Filter1) ++ (0,-\deltalayer) node [filter]       (L2 Filter1) {\two};
  \path (L2 Filter1) ++ \deltasigma      node [nonlinearity] (L2 F1)      {\sigmatwo};
  \path (L2 F1) ++ \deltasigma node [pooling] (L2 F2)      {\textcolor{black}{\proytwo}};
  \path[draw, -stealth] (L2 Filter1.east) --  node [above] {$\mathbf{y}_2$} (L2 F1.west);
  \path[draw, -stealth] (L2 F1.east) -- node [above] {$\mathbf{z}_2$} (L2 F2.west);  
  
  \path (L2 Filter1) ++ (0,-\deltalayer) node [filter]       (L3 Filter1) {\three};
  \path (L3 Filter1) ++ \deltasigma      node [nonlinearity] (L3 F1)      {\sigmathree};
  \path (L3 F1) ++ \deltasigma node [pooling] (L3 F2)      {\textcolor{black}{\proythree}};
  \path[draw, -stealth] (L3 Filter1.east) --  node [above] {$\mathbf{y}_3$} (L3 F1.west);
  \path[draw, -stealth] (L3 F1.east) -- node [above] {$\mathbf{z}_3$} (L3 F2.west); 

  \path[draw, -stealth] (input.east) -- (L1 Filter1.north);
  \path (L1 F2.south) ++ (0,-\deltaconnector) node [] (aux1) {};
  \path[draw, -stealth] (L1 F2.south) -- node [below right] {$\mathbf{x}_1$} (aux1.north) 
                                      --                         (aux1.north -| L2 Filter1.north) 
                                      -- node [above left]  {$\mathbf{x}_1$} (L2 Filter1.north);
  \path (L2 F2.south) ++ (0,-\deltaconnector) node [] (aux1) {};
  \path[draw, -stealth] (L2 F2.south) -- node [below right] {$\mathbf{x}_2$} (aux1.north) 
                                      --                         (aux1.north -| L2 Filter1.north) 
                                      -- node [above left]  {$\mathbf{x}_2$} (L3 Filter1.north);
  \path[draw, -stealth] (L3 F2.south) -- ++ \deltaoutput -- ++ (0.5, 0) 
                        node [right]{$\mathbf{x}_3$};

  \begin{pgfonlayer}{bg} 
      \path (L1 Filter1.west |- L1 F1.south) ++ (-0.4,-0.7)
           node [filter, anchor = south west,
                 fill = black!5, 
                 minimum width  = 11.8*\unit,
                 minimum height = 2.4*\unit,] 
        (layer)
        {}; 
       \path (layer.south west) ++ (0.0,0.0) node [above right] {$(\ccalA_1, \ccalM_1, \rho_1)$};
      \path (L1 Filter1.west |- L2 F1.south) ++ (-0.4,-0.7)
           node [filter, anchor = south west,
                 fill = black!5, 
                 minimum width  = 11.8*\unit,
                 minimum height = 2.4*\unit,] 
        (layer)
        {}; 
       \path (layer.south west) ++ (0.0,0.0) node [above right] {$(\ccalA_2, \ccalM_2, \rho_2)$};
      \path (L1 Filter1.west |- L3 F1.south)  ++ (-0.4,-0.7)
           node [filter, anchor = south west,
                 fill = black!5, 
                 minimum width  = 11.8*\unit,
                 minimum height = 2.4*\unit,] 
        (layer)
        {}; 
       \path (layer.south west) ++ (0.0,0.0) node [above right] {$(\ccalA_3, \ccalM_3, \rho_3)$};  \end{pgfonlayer}

\end{tikzpicture}} 

%% file: v25/perturbations_and_stability.tex




\section{Algebraic Perturbation Models}\label{sec:perturbandstability}

For our discussion we consider perturbations of the generic algebraic signal model $(\ccalA,\ccalM,\rho)$ determined by a perturbation of $\rho$. As discussed in~\cite{parada_algnn}, if the realization of algebraic filters in the algebra is achieved by $\rho$, it is natural to consider that mismatches in the model occur on $\rho$. In the following definition we state formally the notion of perturbation in the context of ASP.


\begin{definition}\label{def:perturbmodel}(ASP Model Perturbation~\cite{parada_algnn})
Let $(\ccalA,\ccalM,\rho)$ be an ASP model with algebra elements generated by $g\in\ccalG$ (Definition \ref{def_generators}) and recall the definition of the shift operators $\bbS = \rho(g)$ (Definition \ref{def_shift_operators}). We say that $(\ccalA,\ccalM,\tdrho)$ is a perturbed ASP model if for all $a=p(\ccalG)$ we have that
\begin{equation}\label{eqn_def_perturbation_model_10}
   \tdrho(a) = p\big(\tdrho(g)\big) 
             =p\big(\tilde\ccalS\big),
\end{equation}
where $\tilde\ccalS$ is a set of perturbed shift operators of the form
\begin{equation}\label{eqn_def_perturbation_model_20}
   \tbS = \bbS + \bbT(\bbS),
\end{equation}
for all shift operators $\bbS\in\ccalS$. 
\end{definition}



From Definition~\ref{def:perturbmodel} we can see that the effect of a perturbation on the homomorphism $\rho$ is expressed in terms of perturbations of the shift operators, which at the same time produce a perturbed version of any algebraic filter. Notice that $\tdrho$ is a general map that is not necessarily a homomorphism, but it could be. As shown in~\cite{parada_algnn} this notion of perturbation is associated to practical scenarios in GSP, WSP and group signal processing.

The notion of stability is tied to the concept of size of a deformation. To measure the size of those deformations on the space of operators we use norms. In particular, we use the operator norms induced by a norm associated to $\ccalM$. If  $\text{dim}(\ccalM)<\infty$ the specifics of the norm in $\ccalM$ are not relevant since all norms are equivalent~\cite{conway1994course}. If $\ccalM$ is infinite dimensional we select a norm that guarantees that every bounded operator in $\text{End}(\ccalM)$ is Hilbert-Schmidt and that the induced norm in a direct sum of replicas of $\ccalM$ satisfies the maximum property -- see Appendix~\ref{app_normsofdirectsums} in the supplementary material. We remark that this does not affect the structure of any algebraic signal model $(\ccalA, \ccalM, \rho)$. We now state the formal definition of stability considered in our analysis.


\begin{definition}[Operator Stability~\cite{parada_algnn}]\label{def:stabilityoperators1} Given operators $p(\mathbf{S})$ and $p(\tilde{\mathbf{S}})$ defined on the models $(\ccalA,\ccalM,\rho)$ and $(\ccalA,\ccalM,\tdrho)$ (cf. Definition \ref{def:perturbmodel}) we say the operator $p(\mathbf{S})$ is Lipschitz stable if there exist constants $C_{0}, C_{1}>0$ such that 
\begin{multline}\label{eq:stabilityoperators1}
\left\Vert p(\mathbf{S})\mathbf{x}  - p(\tilde{\mathbf{S}})\mathbf{x}\right\Vert
\leq
\\
\left[
C_{0} \sup_{\bbS\in\ccalS}\Vert\mathbf{T}(\mathbf{S})\Vert + C_{1}\sup_{\bbS\in\ccalS}\big\|D_{\bbT}(\bbS)\big\|
+\mathcal{O}\left(\Vert\mathbf{T}(\mathbf{S})\Vert^{2}\right)
\right] \big\| \bbx \big\|,
\end{multline}
for all $\bbx\in\ccalM$. In \eqref{eq:stabilityoperators1} $D_{\bbT}(\bbS)$ is the Fr\'echet derivative of the perturbation operator $\bbT$. 
\end{definition}





The right hand side of~(\ref{eq:stabilityoperators1}) provides a measure of the deformation produced by $\bbT(\bbS)$. Then, Definition~\ref{def:stabilityoperators1} states that a given operator is stable to a perturbation $\bbT(\bbS)$ if the deformation induced in the operator is proportional to the size of the deformation. For our discussion we consider the perturbation model given by
\begin{equation}\label{eqn_perturbation_model_absolute_plus_relative}
\bbT(\bbS_i)=\bbT_{0,i} + \bbT_{1,i}\bbS_i,
\end{equation}
which is composed of an absolute or additive perturbation $\bbT_{i,0}$ and a relative or multiplicative perturbation $\bbT_{i,1}\bbS_i$. The family of perturbations is ruled by the condition
\begin{equation}
\left\Vert 
\bbT_{i,r}
\right\Vert_{F}
\leq
\delta
\left\Vert 
\bbT_{i,r}
\right\Vert 
,                
\end{equation}
where $\delta>0$. This is, the Frobenius norm of the perturbation operators is bounded by a scalar factor of the operator norm.

\begin{remark}\normalfont

The notion of stability discussed here is stability to deformations on the \emph{operator} space. This is the same kind of deformation that is studied in \cite{parada_algnn} for commutative algebraic filters and in \cite{gama2020stability} for graph filters. It is not the same as the \emph{domain} deformations studied in \cite{mallat_ginvscatt}. Although different in principle, both notions can be related~\cite{parada_algnn}. For instance, a small deformation of the time axis for time signals implies that signals will not have translation symmetry but instead quasi translation symmetry. This can be seen as a perturbation of the time delay operator. It is important to point out that in all scenarios -- domain and operator deformations -- a stable operators must have changes that are proportional in size to the given deformations. This is done by using metrics used to measure the size of the diffeomorphisms involved in each case \cite{hirsch1997differential, LESLIE1967263, banyaga2013structure}. 

\end{remark}




\section{Stability Theorems}
\label{sec:stabilitytheorems}

In the following definitions we state properties of algebraic filters in $\ccalA$ in terms of their spectral representations in $\tilde{\ccalA}$ (see Def.~\ref{definition:filtspecrep}). This will be used in the derivation of the stability results. We start introducing the notions of Lipschitz and integral Lipschitz filters.


\begin{definition}\label{def_lipschitz_intlipschitz}
Let $(\ccalA, \ccalM, \rho)$ be an ASM where $\ccalA$ has generators $\{ g_i \}_{i=1}^{m}$. Let $\tilde{\ccalA}$ the algebra of matrices containing the spectral representations of the elements in $\ccalA$ (see Def.~\ref{definition:filtspecrep}). We say that $p(g_1,\ldots,g_m)\in\ccalA$ is $L_0$-Lipschitz if there exists $L_{0}>0$ such that
\begin{multline}
\left\Vert
p(x_{1},\ldots,x_{m})-p(\tilde{x}_{1},\ldots,\tilde{x}_{m})
\right\Vert  
\\
\leq
L_{0}\left\Vert 
(x_{1},\ldots,x_{m})-(\tilde{x}_{1},\ldots,\tilde{x}_{m})
\right\Vert,                
\end{multline}
for all $x_{i}, \tilde{x}_{i}$ and where $p(x_{1},\ldots,x_{m})\in\tilde{\ccalA}$. Additionally, it is said that $p(g_1,\ldots,g_m)\in\ccalA$ is $L_1$-integral Lipschitz if there exists $L_{1}>0$ such that
\begin{equation}
\left\Vert 
\bbD_{p\vert x_{i}}(x_{1},\ldots,x_{m})
\left\lbrace \left(\cdot\right)x_{i}\right\rbrace   
\right\Vert
\leq L_{1}
,
\end{equation}
$\forall~x_{i}$, where $\bbD_{p\vert x_{i}}(x_{1},\ldots,x_{m})$ is the partial Fr\'echet derivative of $p(x_{1},\ldots,x_{m})\in\tilde{\ccalA}$, and $\Vert \cdot\Vert$ is the operator norm.
 
\end{definition}


From now on we denote the set of algebraic Lipschitz filters by $\ccalA_{L_{0}}$ and the set of algebraic integral Lipschitz filters by $\ccalA_{L_{1}}$. 

Now we introduce the first stability theorem for algebraic signal models with non commutative algebras with multiple generators.


\begin{theorem}\label{theorem:HvsFrechetmultgen}
Let $(\mathcal{A},\ccalM,\rho)$ be an ASM where $\ccalA$ is generated by $\{g_{i}\}_{i=1}^{m}$ and let $\rho(g_{i})=\bbS_{i}\in\text{End}(\mathcal{M})$ for all $i$. Let $\tilde{\rho}(g_{i})=\tilde{\bbS}_{i}\in\text{End}(\mathcal{M})$, where $(\ccalA, \mathcal{M},\tilde{\rho})$ is a perturbed version of $(\ccalA, \mathcal{M},\rho)$ and $\tilde{\bbS}_{i}$ is related with $\bbS_{i}$ by the perturbation model in~(\ref{eqn_def_perturbation_model_20}). Then, for any $p\in\mathcal{A}$ we have
\begin{multline}
\left\Vert 
               p(\bbS_1,\ldots,\bbS_m)\mathbf{x}-p(\tilde{\bbS}_1,\ldots,\tilde{\bbS}_m)\mathbf{x}
\right\Vert
                 \leq 
\\
\Vert\mathbf{x}\Vert
\sum_{i=1}^{m}
                  \left(
                        \left\Vert D_{p\vert\mathbf{S}_{i}}(\bbS_1,\ldots,\bbS_m)\mathbf{T}(\mathbf{S}_{i})\right\Vert+\mathcal{O}\left(\Vert\mathbf{T}(\mathbf{S}_{i})\Vert^{2}\right)
                  \right)
                  ,
\label{eq:HSoptboundmultgen}
\end{multline}
where $D_{p\vert\mathbf{S}_{i}}(\mathbf{S})$ is the partial Fr\'echet derivative of $p$ on $\mathbf{S}_{i}$.
\end{theorem}

\begin{proof}
	See Appendix~\ref{prooftheoremHvsFrechetmg}.
\end{proof}


It is worth pointing out that in~(\ref{eq:HSoptboundmultgen}) the upper bound adds the contributions of the deformation in each direction $\bbS_{i}$ given by the shift operators. Each individual contribution associated to the perturbation of $\bbS_i$ is determined by the Fr\'echet derivative of the filters acting on the perturbation, and this is true no matter what function $\bbT(\bbS_i)$ is being considered. 

In the following theorems we provide the basic stability result for algebraic filters showing how with a restriction of the filters shaped by the functional form of the right hand side of~\eqref{eq:HSoptboundmultgen} leads to stability.


\begin{theorem}\label{theorem:uppboundDHmultg}
	
Let $(\mathcal{A},\ccalM,\rho)$ be an ASM where $\ccalA$ is a non commutative algebra with $m$ generators $\{g_{i}\}_{i=1}^{m}$. Let $(\ccalA,\mathcal{M},\tilde{\rho})$ a perturbed version of $(\ccalA,\mathcal{M},\rho)$ by means of the perturbation model in~(\ref{eqn_perturbation_model_absolute_plus_relative}). Then, if $p\in\mathcal{A}_{L_{0}}\cap\mathcal{A}_{L_{1}}$ it holds that
\begin{multline}
\left\Vert
             D_{p\vert\mathbf{S}_{i}}(\bbS_1,\ldots,\bbS_m)\bbT(\bbS_{i})
\right\Vert
                \leq
                      \delta L_{0}\sup_{\bbS_{i}\in\mathcal{S}}
                                       \Vert
                                               \bbT(\bbS_{i})
                                      \Vert 
                                      \\
                                      +
                      \delta L_{1}\sup_{\bbS_{i}\in\mathcal{S}}
                                       \Vert 
                                               D_{\bbT}(\bbS_{i})
                                       \Vert
                                       .
\label{eq:DHTSmultg}
\end{multline}
 \end{theorem}
 \begin{proof}  See Appendix~\ref{proof_theorem_DHvsDTmg}.  \end{proof}
 

From Theorems~\ref{theorem:HvsFrechetmultgen} and~\ref{theorem:uppboundDHmultg} we can state the stability results for filters in algebraic models with multiple generators in the following corollary.


\begin{corollary}\label{corollary:stabmultgenfilt}
	Let $(\ccalA,\ccalM,\rho)$ be a non commutative ASM where $\ccalA$ has generators $\{ g_{i}\}_{i=1}^{m}$. Let $(\ccalA,\ccalM,\tilde{\rho})$ be a perturbed version of $(\ccalA,\ccalM,\rho)$ associated to the perturbation model in~(\ref{eqn_perturbation_model_absolute_plus_relative}). If $p\in\mathcal{A}_{L_{0}}\cap\mathcal{A}_{L_{1}}\subset\ccalA$, the operator $p(\bbS_1,\ldots,\bbS_m)$ is stable in the sense of Definition~\ref{def:stabilityoperators1} with $C_{0}=m\delta L_{0}$ and $C_{1}=m\delta L_{1}$.
\end{corollary}
\begin{proof} Replacing~(\ref{eq:DHTSmultg}) from Theorem~\ref{theorem:uppboundDHmultg} into~(\ref{eq:HSoptboundmultgen}) from Theorem~\ref{theorem:HvsFrechetmultgen} and organazing the terms.\end{proof}


Notice that the stability property of the algebraic filters comes at the expense of their selectivity. A restriction on the subsets of the algebra naturally limits the type of frequency representations. Additionally, it is worth pointing out that it is the functional form of $\bbT( \bbS_i )$ the one feature that dictates the types of restrictions imposed on the algebra, necessary to guarantee stability.


\begin{remark}\normalfont
We remark that the non commutativity imposes multidimensional frequency representations, but even in this scenario we have a trade-off between stability and selectivity. We could therefore say that although the non commutativity of the algebra does not change the fact that filters can be stable and selective, the multidimensional nature of the frequency representations leads to restrictions that can be captured only by means of matrix algebras. This in itself implies that two filters with identical functional form -- symbolic expression -- do not have the same stability properties if one of them is associated to a commutative model and the other to a non commutative one. 
\end{remark}



\subsection{Stability of non commutative Algebraic Neural Networks}\label{sec_stability_AlgNN}

In this section we extend the stability results obtained for algebraic filters to operators representing AlgNNs. Let $\left\lbrace \left(\mathcal{A}_{\ell},\mathcal{M}_{\ell},\rho_{\ell}; \sigma_\ell\right) \right\rbrace_{\ell=1}^{L}$ be an AlgNN with $L$ layers, and whose perturbed version is represented as $\left\lbrace \left(\mathcal{A}_{\ell},\mathcal{M}_{\ell},\tilde{\rho}_{\ell};\sigma_\ell\right) \right\rbrace_{\ell=1}^{L}$. We start stating formally  how the algebraic operators are affected by the functions that map information between layers.


\begin{theorem}\label{theorem:stabilityAlgNN0}
	
Let $\left\lbrace\left(\mathcal{A}_{\ell},\mathcal{M}_{\ell},\rho_{\ell};\sigma_\ell \right)\right\rbrace_{\ell=1}^{L}$ be an algebraic neural network and $\left\lbrace\left( \mathcal{A}_{\ell},\mathcal{M}_{\ell},\tilde{\rho}_{\ell} ;\sigma_\ell \right)\right\rbrace_{\ell=1}^{L}$ its perturbed version by means of the perturbation model in~(\ref{eqn_perturbation_model_absolute_plus_relative}). We consider one feature per layer and non commutative algebras $\mathcal{A}_{\ell}$ with $m$ generators. If  $\Phi\left(\mathbf{x}_{\ell-1}, \mathcal{P}_{\ell},\mathcal{S}_{\ell}\right)$ and 
$\Phi\left(\mathbf{x}_{\ell-1},\mathcal{P}_{\ell},\tilde{\mathcal{S}}_{\ell}\right)$ represent the $\ell$-th mapping operators of the AlgNN and its perturbed version, it follows that
\begin{multline}
\left\Vert
               \Phi\left(\mathbf{x}_{\ell-1},\mathcal{P}_{\ell},\mathcal{S}_{\ell}\right)
                -
               \Phi\left(\mathbf{x}_{\ell-1},\mathcal{P}_{\ell},\tilde{\mathcal{S}}_{\ell}\right)
\right\Vert
                 \leq
 \\
                        C_{\ell}\delta
                             \Vert
                                    \mathbf{x}_{\ell-1}
                             \Vert
                             m
                             \left(
                                     L_{0}^{(\ell)} \sup_{\bbS_{i,\ell}}\Vert\bbT^{(\ell)}(\bbS_{i,\ell})\Vert 
                             \right.  
                             \\
                             \left.      
                                     +
                                     L_{1}^{(\ell)}\sup_{\bbS_{i,\ell}}\Vert D_{\mathbf{T^{(\ell)}}}(\bbS_{i,\ell})\Vert
                            \right)
\label{eq:theoremstabilityAlgNN0}
\end{multline}
where $C_{\ell}$ is the Lipschitz constant of $\sigma_{\ell}$, and $\mathcal{P}_{\ell}=\mathcal{A}_{L_{0}}\cap\mathcal{A}_{L_{1}}$ represents the domain of $\rho_{\ell}$. The index $(\ell)$ makes reference to quantities and constants associated to the layer $\ell$.

\end{theorem}
\begin{proof}See Appendix~\ref{prooftheorem:stabilityAlgNN0} in the supplementary material. \end{proof}


Theorem~\ref{theorem:stabilityAlgNN0} highlights the role of the maps $\sigma_\ell$ in the stability of the algebraic operators. In particular, we can see that the effect of such functions is only to scale the stability bound of the algebraic operators. If $C_\ell =1$, the stability bounds for the operators in the layers of the AlgNN are identical to the stability bounds for the algebraic filters. However, it is important to remark that the discriminability of the layer operators in the AlgNN is enriched by the pointwise nonlinearities. 

Notice that the resemblance between the results provided in~Theorem~\ref{theorem:stabilityAlgNN0} and previous results for commutative architectures is rooted in the fact that the same notion of stability and the same type of perturbations are considered in both scenarios. However, there is a substantial difference between the stability bounds. In the non commutative scenario we have that the restrictions on the subsets of the algebra are dictated by matrix functions and not scalar maps. Additionally, as can be seen in the proof of the theorems above, the estimate of the upper bounds on the change of the operators requires a more general representation of spectral responses not given by the traditional spectral theorem.

Now we are ready to state the stability theorem for a general AlgNN.


\begin{theorem}\label{theorem:stabilityAlgNN1}

Let $\left\lbrace\left(\mathcal{A}_{\ell},\mathcal{M}_{\ell},\rho_{\ell};\sigma_\ell \right)\right\rbrace_{\ell=1}^{L}$ be an algebraic neural network and $\left\lbrace\left( \mathcal{A}_{\ell},\mathcal{M}_{\ell},\tilde{\rho}_{\ell} ;\sigma_\ell \right)\right\rbrace_{\ell=1}^{L}$ its perturbed version by means of the perturbation model in~(\ref{eqn_perturbation_model_absolute_plus_relative}). We consider one feature per layer and non commutative algebras $\mathcal{A}_{\ell}$ with $m$ generators. If  $\Phi\left(\mathbf{x},\{ \mathcal{P}_{\ell} \}_{1}^{L},\{ \mathcal{S}_{\ell}\}_{1}^{L}\right)$ and 
$\Phi\left(\mathbf{x},\{ \mathcal{P}_{\ell} \}_{1}^{L},\{ \tilde{\mathcal{S}}_{\ell}\}_{1}^{L}\right)$ represent the mapping operator 
and its perturbed version, it follows that 
\begin{multline}
\left\Vert
\Phi
      \left(
              \mathbf{x},\{ \mathcal{P}_{\ell} \}_{1}^{L},\{ \mathcal{S}_{\ell}\}_{1}^{L}
      \right)
-
\Phi
       \left(
              \mathbf{x},\{ \mathcal{P}_{\ell} \}_{1}^{L},\{ \tilde{\mathcal{S}}_{\ell}\}_{1}^{L}
       \right)
\right\Vert
\\
\leq
\sum_{\ell=1}^{L}\boldsymbol{\Delta}_{\ell}\left(\prod_{r=\ell}^{L}C_{r}\right)\left(\prod_{r=\ell+1}^{L}B_{r}\right)
\left(\prod_{r=1}^{\ell-1}C_{r}B_{r}\right)\left\Vert\mathbf{x}\right\Vert
,
\label{eq:theoremstabilityAlgNN1}
\end{multline}
where $C_{\ell}$ is the Lipschitz constant of $\sigma_{\ell}$, $\Vert\rho_{\ell}(a)\Vert\leq B_{\ell}~\forall~a\in\mathcal{P}_{\ell}$, and $\mathcal{P}_{\ell}=\mathcal{A}_{L_{0}}\cap\mathcal{A}_{L_{1}}$ represents the domain of $\rho_{\ell}$. The functions $\boldsymbol{\Delta}_{\ell}$ are given by
\begin{equation}\label{eq:varepsilonl}
\boldsymbol{\Delta}_{\ell}
           =\delta m\left(
           L_{0}^{(\ell)} \sup_{\bbS_{i,\ell}}\Vert\bbT^{(\ell)}(\bbS_{i,\ell})\Vert 
           \right.
           \left.
          +
          L_{1}^{(\ell)}\sup_{\bbS_{i,\ell}}\Vert D_{\mathbf{T^{(\ell)}}}(\bbS_{i,\ell})\Vert
             \right)
             ,
\end{equation}
with the index $(\ell)$ indicating quantities and constants associated to the layer $\ell$.
\end{theorem}
\begin{proof} See Section~\ref{prooftheorem:stabilityAlgNN1} in the supplementary material.  \end{proof}


This final result highlights that the stability of an AlgNN is inherited from the stability properties associated to the operators in each layer. Each layer of the AlgNN contributes to increase the size of the stability constants, and we can observe that with the appropriate normalization of $C_\ell$ and $B_\ell$, we obtain a stability bound that is in essence the same derived for filters and for the mapping operators in each layer. However, the discriminability power associated to the AlgNN is by far larger than the one related with the operators in the layers and the filters. This is a consequence of the pointwise nonlinearities mapping information between layers. It is important to highlight that the $\sigma_\ell$ that map  information between layers are identical for commutative and non commutative architectures. However, they way they redistribute spectral information is tied to the nature of the spectral representations. While in commutative architectures such redistribution is always done between spaces of the same dimension -- one dimensional representations --, in non commutative AlgNNs the redistribution of spectral information occurs in general between spaces of different dimension.

%% file: v25/sec_numexp.tex


\section{Numerical Experiments}\label{sec_numexp}

%
%
%

In order to provide numerical evidence of the stability results derived for AlgNN with non commutative algebras, we consider two architectures:  Multigraph Neural Networks (MultiGNN) and Quaternion Convolutional Networks (QCN). For the first, 
we propose an architecture for learning on multigraphs and show that if the learned filters are penalized to be Integral
Lipschitz, the model is much more resilient to perturbations enforced on the graph. This architecture is used for a rating prediction task on the MovieLens-100K dataset \cite{harper2015movielens}. We then apply the quaternion architecture introduced in \cite{gaudet2018deep} to an image classification task using the MNIST dataset \cite{deng2012mnist} and observe that model performance quickly diminishes when additive and relative perturbations are applied to the quaternion filters. As we will elaborate in Subsection~\ref{subsec_qcn_num_exp}, due to the cyclic nature of the generators in the quaternion algebra, no filter in~\eqref{eq_algspec_quaternion} can be integral Lipschitz. As a consequence, no filter can be selected to mitigate the perturbations that severely affect the performance of the QCN.



\begin{figure*}[bht]
	\centering
	\begin{subfigure}{.46\textwidth}
		\centering
		\includegraphics[width=1\linewidth]{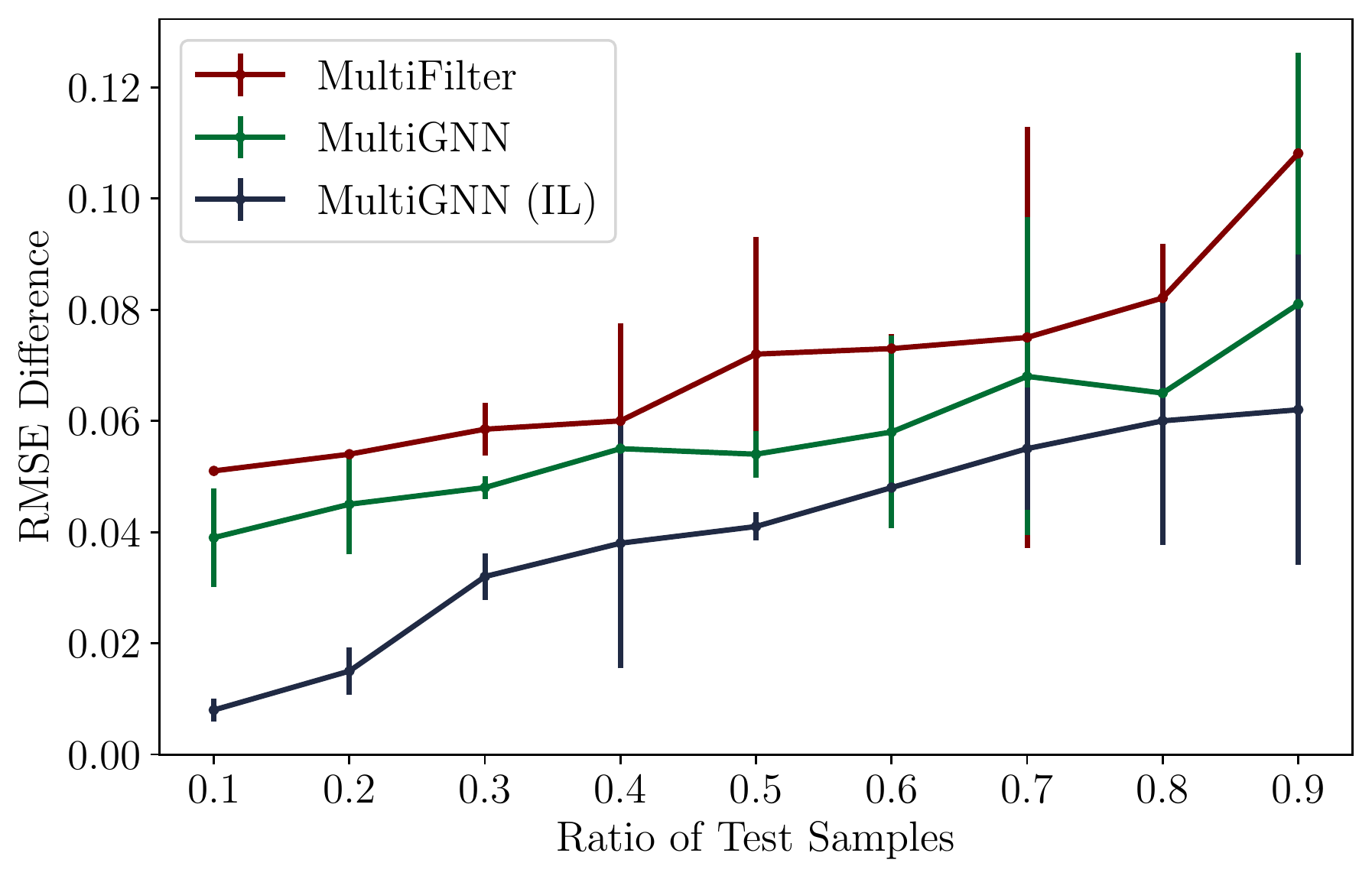} 
		\caption{}
		\label{fig:multiData-one}
	\end{subfigure}
	\begin{subfigure}{.5\textwidth}
		\centering
		\includegraphics[width=1\linewidth]{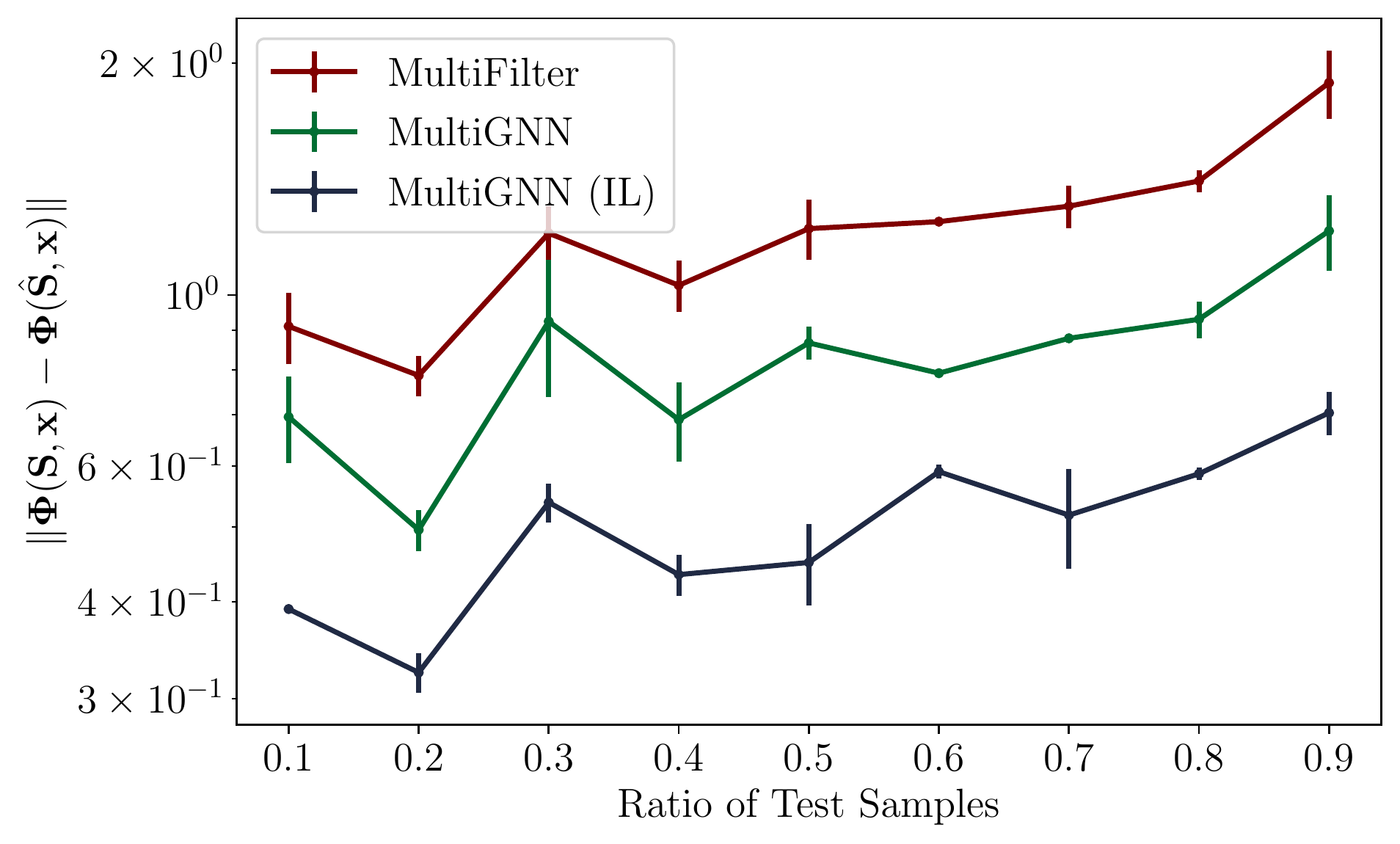} 
		\caption{}
		\label{fig:multiData-two}
	\end{subfigure}
	
	\caption{Stability to perturbations caused by estimation error for the Movie Recommendation problem. We demonstrate in (a) the difference in evaluation measure as the ratio of training samples is increased, where the penalized MultiGNN maintains the smallest difference for all ratios. This trend is also held in (b) where we showcase the difference in the output of the convolutional layer. Then, MultiGNN with IL filters provide more
		stable and consistent rating predictions with respect to estimation error perturbations.}
	\label{fig:multiData}
\end{figure*}


\subsection{Multigraph Neural Networks}

We first consider the application of the Multigraph Neural Network as a movie recommendation system \cite{huang2018rating}. Using the MovieLens-100K dataset containing 100,000 ratings from 943 users, we model the system as a multigraph. Nodes represent movies which are connected by two classes of edges: one measuring the rating similarity and the other measuring the genre similarity. The signals supported by this multigraph are user ratings, with the task of inferring unseen ratings based on ratings given to other movies. On this task, we carry out stability experiments which measure robustness to estimation error from the training set. We find that Multigraph Neural Networks that are penalized to employ integral Lipschitz filters are most stable to perturbations. 

\textbf{Multigraph Formation:} From the 1582 movies available, we use the 200 with the highest number of ratings. The movie \emph{Toy Story} is used as the target movie for predicted ratings. Of users who rated this movie, we define a node signal as the rating 1-5 they gave to each movie or 0 if they did not rate it.  These signals are divided with a 90\%/10\% train/test split. Using the training set, we add rating edge weights between movies using the Pearson correlation coefficient and genre edge weights through the Jaccard similarity of overlapping genres. To appeal to sparsity, the 20 highest weighted edges of both edge classes are kept for each node, and the rest of the edges are removed.

\black{
\textbf{Stability:} In contrast with the scenario of the quaternion algebras, multigraph signal models are associated with a very rich algebra: the regular algebra of polynomials with multiple independent variables. This guarantees the existence of subclasses of filters that are Lipschitz and integral Lipschitz. Therefore, it is possible to learn filters -- Lipschitz and integral Lipschitz -- that can mitigate the effect of deformations on the shift operators. We illustrate this in our experiments analyzing the magnitude of change in the filters and operators networks when subjected to deformations.

}

\textbf{Architectures and Training:} In our stability experiments, we consider three architectures. The first, MultiFilter, is a learned linear multigraph filter. We also train a Multigraph Neural Network (MultiGNN) and another which is regularized by an estimate of the filter’s integral Lipschitz constant (MultiGNN IL), seeking to train filters that are integral Lipschitz. All architectures employ a convolutional layer with 3 filter taps and 64 output features, a ReLU nonlinearity function, and a local linear readout layer mapping the output rating to a single scalar rating estimate. We minimize the smooth L1 loss using an ADAM optimizer for 40 epochs, and evaluate using the root mean squared error (RMSE). Results are reported as the average and standard deviation of performance across 5 random splits of the dataset.


\textbf{Estimation Error Experiment:} To simulate perturbations as a result of estimation error, we first train each architecture on 90\% of the training set. At evaluation, we replace the underlying multigraph support with an estimate generated by a training set ranging from 10\% to 90\% of the size of the overall dataset. First, we find the difference in RMSE between the trained and evaluated models. The regularized MultiGNN maintains the smallest difference for each training set size as demonstrated in Figure \ref{fig:multiData-one}. We also compare the norm of the difference in output of the convolutional layer. As can be seen in Figure \ref{fig:multiData-two}, the MultiGNN with integral Lipschitz filters is approximately an order of magnitude more stable than the unregularized MultiGNN.

\subsection{Quaternion Convolutional Networks}
\label{subsec_qcn_num_exp}

%
%

We now consider a non commutative AlgNN proposed in literature: Quaternion Convolutional Networks. Quaternions have shown to be useful tools for modeling  spatial transformations with diverse applications to computer graphics, quantum mechanics, and signal processing. Gaudet and Maida \cite{gaudet2018deep} propose a deep convolutional architecture for learning through use of quaternion convolutions, quaternion batch-normalization, and a quaternion weight initialization scheme. We use this architecture as the basis for our second stability experiment.

\textbf{Architecture:} Quaternion convolutions occur through convolving a quaternion filter matrix $\mathbf{W} = \mathbf{A} + i \mathbf{B} + j \mathbf{C} + k \mathbf{D}$ with a quaternion vector $\mathbf{h} = \mathbf{w} + i \mathbf{x} + j \mathbf{y} + k \mathbf{z}$ according to~\eqref{eq_ex_quaternionsp_1} -- see Appendix~\ref{sec_quaternions_sp_nn}. In forming a Quaternion Convolutional Network, we compose layers of quaternion convolutions, applying a nonlinear function to the output of each layer. We then append quaternion dense layers, which are traditional fully-connected neural networks applied component-wise. The resultant embedding is shaped into the desired output shape using a readout layer. Our particular architecture is composed of three quaternion convolutional layers (with 32 channels each), two quaternion dense layers (with 82 and 48), and a readout layer, using the ReLU function as the nonlinearity. 

\black{\textbf{Stability:} We recall that one of the requirements to guarantee stability to the perturbations in~\eqref{eqn_perturbation_model_absolute_plus_relative} is the existence of subclasses of filters in~\eqref{eq_algspec_quaternion} that are Lipschitz and integral Lipschitz. Although in principle any filter in~\eqref{eq_algspec_quaternion} can be written by considering arbitrary powers on the generators of the algebra, due to the cyclic nature of $\{ \boldsymbol{1}, \boldsymbol{i}, \boldsymbol{j}, \boldsymbol{k} \}$, such representations reduce to a linear combination of the generators. Therefore, the subclass of integral Lipschitz filters in~\eqref{eq_algspec_quaternion} is empty. Additionally, while the filters in~\eqref{eq_algspec_quaternion} can still be Lipschitz, the Lipschitz constants will be large. Thus, the performance of the QCN architecture -- as defined in~\cite{gaudet2018deep} -- is expected to be severely affected by perturbations, without the potential to design stable filters. We demonstrate this vulnerability in our numerical experiments. 
}


\textbf{Synthetic Experiment:} To corroborate these theoretical results, we first train an unperturbed architecture. Then, during evaluation, we inject noise in the first quaternion convolutional layer by perturbing $\mathbf{A}$, a layer of the quaternion filter matrix $\mathbf{W}_1$. \black{More precisely, we first consider additive perturbations by replacing $\mathbf{A}$ with $\Tilde{\mathbf{A}}_1 = \mathbf{A} + \mathbf{T}_1$, where $\mathbf{T}_1$ has its components sampled from the uniform distribution $[-\varepsilon_1, \varepsilon_1]$. We also observe relative perturbations by substituting $\mathbf{A}$ with $\Tilde{\mathbf{A}}_2 = \mathbf{A} + \mathbf{T}_2\mathbf{A}$, where $\mathbf{T}_2$ has its values drawn from the uniform distribution $[\varepsilon_2, \varepsilon_2]$.}


\setlength{\tabcolsep}{3.4pt}
\begin{table}[t]
\black{
\centering
\begin{tabular}{c|ccc|c|ccc}
\hline
\hline
              & \multicolumn{3}{c|}{Additive Perturbation} & & \multicolumn{3}{c}{Relative Perturbation} \\
$\varepsilon_1$ & First Con.     & Last Con.    & Acc.  &  $\varepsilon_2$  & First Con.      & Last Con.      & Acc.       \\ \hline
0.005          & 27,813          & 4,415       & 0.991    &  0.05 & 24,986              & 6,316             & 0.991      \\
0.01          & 55,467         & 9,542      & 0.991    & 0.1 & 49,961            & 13,100           & 0.990      \\
0.05           & 289,610         & 78,092       & 0.968    & 0.5 & 256,754            & 79,819           & 0.956      \\
0.1           & 604,553       & 213,046       & 0.772    & 1& 533,462           & 158,565          & 0.825     \\ 
0.5           & 3,303,611       & 234,771       & 0.222    & 5 & 2,975,365           & 1,361,751          & 0.321      \\ \hline
\end{tabular}
}

	\caption*{\small{Table 1. Stability of Quaternion Convolutional Networks to additive and relative perturbation.}}
    \label{table:simulationsQuat}
	
\end{table}

We apply this architecture to the image classification task using the MNIST dataset \cite{deng2012mnist}. \black{Following \cite{parcollet2019quaternion}, we form the quaternion input for each pixel by repeating the grayscale value along each of the three imaginary components and zeroing out the real component.} We train the unperturbed architecture for 10 epochs using a batch size of 500. After training, we first evaluate the performance of the unperturbed architecture on the test samples. We then apply additive and multiplicative perturbation to the first quaternion convolutional layer for various values of $\varepsilon_1$ and $\varepsilon_2$, and observe the new accuracy (Acc.) and norm of the difference between the perturbed and unperturbed architectures of the output from the first quaternion convolutional layer (First Con.) and the last quaternion convolutional layer (Last Con.). Average results across 10 noise samplings are reported in Table 1.

Our results find that Quaternion Convolutional Networks are particularly sensitive to small additive and relative perturbations  applied in the first layer, as performance quickly diminishes as $\varepsilon_1$ and $\varepsilon_2$ increases. These results indicate that an architecture trained to strictly minimize loss not result in stable filters.

%% file: v25/sec_discussion.tex


\section{Discussion and Conclusions}\label{sec:discussion}

We analyzed non commutative algebraic neural networks (AlgNNs) providing general stability results applicable to convolutional architectures such as multigraph-CNNs, quaternion-CNNs, quiver-CNNs, and Group-NNs (non commutative) among others. We have shown that non commutative convolutional neural networks can be stable. This property has an analogous form to the stability bounds of commutative signal models but with substantial differences. As we showed in previous sections, the restrictions necessary to guarantee stability are defined by algebras of matrices, which highlights the fact that the stability of filters in non commutative signal models is not a trivial extension of the stability results for commutative models. This is a consequence of the fact that spectral representations in non commutative signal models are given by matrix and not scalar functions.

As shown in Theorem~\ref{theorem:HvsFrechetmultgen}, the size of the deformation of a filter operator $p(\bbS_1,\ldots, \bbS_m)$ is bounded by the norm of the Fr\'echet derivative of $p$  acting on $\bbT(\bbS_i)$. This holds for \textit{any arbitrary perturbation} $\bbT(\bbS_i)$ indicated in~(\ref{eqn_def_perturbation_model_20}) and it highlights the versatility of the deformation model for the representation of a large variety of perturbations.

Notice that non commutative models naturally embed properties of those signal frameworks where there are not shift operator invariances, i.e. $\bbS_i p(\bbS_1,\ldots,\bbS_m)\neq p(\bbS_1, \ldots, \bbS_m) \bbS_i$. The stability results derived in this paper show that despite the lack of this invariance, stability is possible and it comes at a price on selectivity and with stronger restrictions on the spectral responses of the filters. 

We point out that our results can be extended to more elaborated versions of $\bbT(\bbS_i)$. This opens up interesting research applications where $\bbT(\bbS_i)$ can be tailored to more specific scenarios where stable filters could be designed for $\bbT (\bbS_i) = \sum_{r}\bbT_r \bbS_{i}^r$, and the restrictions imposed on the algebra will be determined by the properties of the operators $\bbT_r$.

%% file: v25/sec_basic_algebra.tex


\section{Basic Algebra Concepts}
\label{sec_basic_algebra}

\textbf{Algebra}: We say that the vector space $\ccalA$ is an algebra if $\ccalA$ is equipped with a product. This is, there is an operation of product between two elements of $\ccalA$ that results in another element of $\ccalA$~\cite{repthybigbook}. Let us denote by $ab$ the product between $a$ and $b$ with $a,b\in\ccalA$. Then, we say that the algebra $\ccalA$ is unital if there exists a multiplicative unit element $\mathbf{1}\in\ccalA$ such that $a\mathbf{1} = \mathbf{1} a = a$ for all $a\in\ccalA$. Typical examples of algebras are:

\begin{itemize}
\item $\mathbb{C}$: The set of complex numbers is a vector space over $\mathbb{C}$ itself. Using as algebra product in $\mathbb{C}$ the ordinary product between complex numbers we can see $\mathbb{C}$ as an algebra.
\item $\mathbb{C}[t]$: The set of polynomials with coefficients in $\mathbb{C}$ and independent variable $t$ is a vector space over $\mathbb{C}$. Using as algebra product the ordinary product between polynomials we can see $\mathbb{C}[t]$ as an algebra. 
\item $M_{r\times r}(\mathbb{C})$: The space of matrices of size $r\times r$ with entries in $\mathbb{C}$ is a vector space over $\mathbb{C}$. Using as algebra product the ordinary product between matrices we can see $M_{r\times r}(\mathbb{C})$ as an algebra.
\end{itemize}

\noindent It is important to remark that the formal definition of a \textit{product} is given by a bilinear map. We say that the map $m_{\ccalA} : \ccalA \times \ccalA \to \ccalA$ is bilinear if $m_{\ccalA}$ is linear on each argument. Then, talking about an operation that behaves like a product we talk about a bilinear map that assigns the value of the given product~\cite{repthybigbook}.

\textbf{Endomorphisms}: Let $\ccalM$ be a vector space and let $\text{End}(\ccalM)$ be the space of linear maps from $\ccalM$ onto itself. If $\text{dim}(\ccalM) = n$, then $\text{End}(\ccalM)$ is isomorphic to the space of matrices of size $n\times n$. This is, whenever the dimension of $\ccalM$ is finite we can think about $\text{End}(\ccalM)$ as a space of matrices.

%% file: v25/sec_group_sp_nn.tex


\section{Group Signal Processing and Group Neural Networks}
\label{sec_grsp_discussion}


		
Classical convolutional information processing on locally compact groups is given by the algebraic model $(\ccalA_G,\ccalM,\rho)$, where
$
\ccalM =
         \left\lbrace
                     \sum_{g\in G}\bbx(g)g,\quad\bbx(g)\in\mathbb{C}
         \right\rbrace
$
is the set of complex valued functions defined on the group $G$. The algebra $\ccalA_G$ is the \textit{group algebra}, which is given by $\ccalA_G=\ccalM$. The homomorphism $\rho$ is given by $\rho(\bba)=L_{\bba}$ with $L_{\bba}\bbx=\bba\bbx$. Then, the convolution between a group filter and a group signal takes the form
\begin{equation}\label{eqn_group_filters}
\rho\left(\sum_{g\in G}\bba(g)g\right)\bbx=\sum_{g\in G}\bba(g)g\bbx=\sum_{g\in G}\sum_{h\in G}\bba(g)\bbx(h)gh.
\end{equation}
Notice that the representation of the convolution can also be re written making $u=gh$, which leads to~\cite{terrasFG}
\begin{equation}\label{eqn_group_convolution}
   \sum_{g,h\in G} \bba(g)\bbx(h)gh 
             =\sum_{u,h\in G}\bba(uh^{-1})\bbx(h)u .
\end{equation}
%
%
%


We emphasize that the role of this example is to show that classical group convolutions -- as considered in~\cite{Bronstein2021GeometricDL} -- are \textit{one} particular case of algebraic convolutions. What is more, it is one particular case of the many algebraic convolutions that can be associated to the same group algebra. This is, given a non commutative group $G$ and its associated group algebra $\ccalA_G$, it is possible to define different convolutional models $(\ccalA_G, \ccalM,\rho)$ when different choices of $\ccalM$ and $\rho$ are made~\cite{Kumar2022AlgebraicCF}. This implies that it is possible to leverage the symmetries of the group under a convolutional signal model in different ways and in vector spaces of arbitrary dimension.

Notice that when convolutions are computed as in~\cite{Bronstein2021GeometricDL}, a \textit{lifting} process is required to map the signals -- usually not defined on the group -- on the group $G$. If $G$ is continuous  -- a Lie group -- computing the group convolution requires the approximation of a Haar integral that generalizes~\eqref{eqn_group_convolution} to locally compact groups~\cite{folland2016course}. In~\cite{Kumar2022AlgebraicCF} the computing of the convolutions -- including convolutions with Lie groups -- is performed considering an ASM $(\ccalA_G, \ccalM, \rho)$, where  $\ccalA_G$ is the group algebra associated to the group $G$, $\ccalM$ is a vector space of arbitrary dimension and $\rho$ is the homomorphism that will translate approximations of elements in $\ccalA_G$ into concrete operators. In~\cite{Kumar2022AlgebraicCF} lifting is not necessary since the implementation of the filters in $\ccalA_G$ is carried out by means of an efficient approximation of $\rho$.

A group convolutional neural network is a stacked layered structure where information in each layer is processed using group convolutions. With convolutions performed as in~\cite{Kumar2022AlgebraicCF}, in each layer of the architecture information is processed according to the triplet $(\ccalA_G, \ccalM_\ell, \rho_\ell)$. The information is mapped between layers by means of $\sigma_\ell = P_\ell \eta_\ell$ where $\eta_\ell$ is a pointwise nonlinearity and $P_\ell$ is a dimensionality reduction operator. When dealing with discrete scenarios we have $\text{dim}(\ccalM)<\infty$ and the action of $\eta_\ell$ is expressed in terms of a canonical basis in $\mathbb{R}^{\text{dim}(\ccalM)}$.

Notice that the processing of information with ~\cite{Bronstein2021GeometricDL} can be seen as a particular case of the convolutional architectures proposed in~\cite{Kumar2022AlgebraicCF} with the advantage that in~\cite{Kumar2022AlgebraicCF} there is no need for the discretization of the Haar integral since the implementation of the filters relies on building an accurate approximation of $\rho_\ell$ in $(\ccalA_G, \ccalM_\ell, \rho_\ell)$.

%% file: v25/sec_quaternion_sp_nn.tex


\section{Quaternion Signal Processing and Quaternion Neural Networks}
\label{sec_quaternions_sp_nn}


	
Let $\ccalQ$ be the algebra on $\mbC$ generated by the symbols $\boldsymbol{i}, \boldsymbol{j}, \boldsymbol{k}$ and the unit element $\mathbf{1}$, where
%
%
%
$
\boldsymbol{i}^2 = \boldsymbol{j}^2 = \boldsymbol{k}^2 =-\mathbf{1}
,
\quad 
\boldsymbol{ij} = -\boldsymbol{ji} = \boldsymbol{k}
,
\quad
%
%
%
\boldsymbol{jk} = -\boldsymbol{kj} = \boldsymbol{i}
,
\quad
\boldsymbol{ki} = -\boldsymbol{ik} = \boldsymbol{j}
.
$
%
%
%
$\ccalQ$ is known as the algebra of quaternions~\cite{repthysmbook}. Using $\ccalQ$ is possible to build quaternion-based algebras that can be used to define convolutional models. In~\cite{gaudet2018deep} such models are build upon the algebra
\begin{equation}\label{eq_algspec_quaternion}
\ccalA = \left\lbrace\left.
                             p_1 (t)\boldsymbol{1} + p_2 (t)\boldsymbol{i} + p_3 (t)\boldsymbol{j}
                             + p_4 (t)\boldsymbol{k} 
                             \right\vert                              \boldsymbol{1},\boldsymbol{i},\boldsymbol{j},\boldsymbol{k}\in\ccalQ   
                 \right\rbrace
                 ,
\end{equation}
and where $p_i (t) = \sum_{k=0}^{k}h_{i,k}t^k$ are elements of the algebra of polynomials of a single variable. We can see that $\ccalA$ is a vector space when considering the sum component wise on the quaternion elements, and the multiplication by scalars. The product algebra in $\ccalA$ follows the distribution product of the quaternions. Then, we can define the quaternion convolutional model $(\ccalA, \ccalM, \rho)$, where 
\begin{equation}
\ccalM =  \left\lbrace\left.
                    \bbx_1\boldsymbol{1} + \bbx_2\boldsymbol{i} + \bbx_3\boldsymbol{j}
                                              + \bbx_4\boldsymbol{k} 
                      \right\vert 
                           \boldsymbol{1},\boldsymbol{i},\boldsymbol{j},\boldsymbol{k}\in\ccalQ;
                      \quad
                      \bbx_i\in\mbR^{N}
                      \right\rbrace
,
\end{equation} 
while the homomorphism $\rho$ is given according to
\begin{multline}
\rho\left(  
          p_1 (t)\boldsymbol{1} + p_2 (t)\boldsymbol{i} + p_3 (t)\boldsymbol{j}
          + p_4 (t)\boldsymbol{k} 
    \right)
    \\
    =
    p_1 (\bbC)\boldsymbol{1} + p_2 (\bbC)\boldsymbol{i} + p_3 (\bbC)\boldsymbol{j}
    + p_4 (\bbC)\boldsymbol{k} 
    ,
\end{multline}
where $\rho(t)=\bbC\in\mbR^{N\times N}$ is the discrete time delay operator. Then, the convolutional action of a filter $\rho(f)=\bbF$ on $\boldsymbol{u}\in\ccalM$ is given by
\begin{multline}\label{eq_ex_quaternionsp_1}
\bbF\boldsymbol{u}
=
\left( p_{1}(\bbC)\bbx_1 -p_{2}(\bbC)\bbx_2 - p_{3}(\bbC)\bbx_3-p_{4}(\bbC)\bbx_4 \right)\boldsymbol{1}+
\\
\left( p_{1}(\bbC)\bbx_2 -p_{2}(\bbC)\bbx_1 - p_{3}(\bbC)\bbx_4-p_{4}(\bbC)\bbx_3 \right)\boldsymbol{i} +
\\  
\left( p_{1}(\bbC)\bbx_3 -p_{2}(\bbC)\bbx_4 - p_{3}(\bbC)\bbx_1-p_{4}(\bbC)\bbx_2 \right)\boldsymbol{j}+
\\
\left( p_{1}(\bbC)\bbx_4 -p_{2}(\bbC)\bbx_3 - p_{3}(\bbC)\bbx_2-p_{4}(\bbC)\bbx_1 \right)\boldsymbol{k}
.
\end{multline}
Notice that $p_{i}(\bbC)\bbx_i$ is the ordinary Euclidean convolution in $\mbR^N$ between a filter and the signal $\bbx_i$. \black{We emphasize that the particular form of~\eqref{eq_ex_quaternionsp_1} is a consequence of the non commutativity of the filter operators from $\ccalA$ in~\eqref{eq_algspec_quaternion}, which is inherited from the quaternion algebra.}
%
%
%

A quaternion neural network as introduced in~\cite{gaudet2018deep} is a stacked layered structure where information is processed in each layer according to~\eqref{eq_ex_quaternionsp_1}. The information processed in the layer $\ell$ has the form $\boldsymbol{u}_\ell = \bbx_1\boldsymbol{1} + \boldsymbol{i}\bbx_2 + \boldsymbol{j}\bbx_3 + \boldsymbol{k}\bbx_4$, while the convolutional filters are of the form $\bbF = \bbA\boldsymbol{1} + \boldsymbol{i}\bbB + \boldsymbol{j}\bbC + \boldsymbol{k}\bbD$. The action of the filter $\bbF$ on the signal $\boldsymbol{u}$ follows the ordinary product between quaternions considering that $\bbA,\bbB,\bbC,\bbD$ act as Euclidean convolution operators on $\bbx_i$ as indicated in~\eqref{eq_ex_quaternionsp_1}. Then, given an input $\boldsymbol{u}_\ell$ to the $\ell$-th layer we leverage the symmetries given by the quaternion algebra to obtain $\bby_\ell = \bbF \boldsymbol{u}_\ell$. After this a point-wise nonlinear operator $\eta_\ell$ is applied to obtain $\bbz_\ell = \eta_\ell (\bby_\ell)$ with $\bbz_\ell (u) = \max\left\lbrace \bby_\ell (u), 0\right\rbrace$, and where $\bbz_\ell (u)$ is the $u$-th component of $\bbz_\ell$. A conventional pooling operator $P_\ell:\mbR^N\rightarrow\mbR^M$ performs traditional sampling on a quaternion signal according to $P_{\ell}(\boldsymbol{u}_\ell )=\boldsymbol{1}P_{\ell}(\bbx_1)+ \boldsymbol{i}P_{\ell}(\bbx_2)+\boldsymbol{j}P_{\ell}(\bbx_3)+\boldsymbol{k}P_{\ell}(\bbx_4)$.

%% file: v25/sec_proofTheorems.tex

\section{Proof of Theorems}\label{sec:proofofTheorems}

\subsection{Proof of Theorem~\ref{theorem:HvsFrechetmultgen}}
\label{prooftheoremHvsFrechetmg}

\begin{proof}
	To simplify notation we use  $\bbS=(\bbS_1,\ldots,\bbS_m)$, and we start taking into account the definition of the Fr\'echet derivative	
	\begin{equation}
	p(\mathbf{S}+\boldsymbol{\xi})=p(\mathbf{S})+D_{p}(\mathbf{S})\left\lbrace\boldsymbol{\xi}\right\rbrace+o(\Vert\boldsymbol{\xi}\Vert).
	\label{eq:frechetDdefmg}
	\end{equation}
	Taking the norm on both sides of~(\ref{eq:frechetDdefmg}) and applying the triangle inequality it follows that
	\begin{equation}
	\left\Vert p(\mathbf{S}+\boldsymbol{\xi})-
	p(\mathbf{S})\right\Vert\leq
	\left\Vert D_{p}(\mathbf{S})\left\lbrace\boldsymbol{\xi}\right\rbrace\right\Vert+\mathcal{O}\left(\Vert\boldsymbol{\xi}\Vert^{2}\right)    
	\end{equation}
	for all $\boldsymbol{\xi}=(\boldsymbol{\xi}_{1},\ldots,\boldsymbol{\xi}_{m})\in\text{End}(\mathcal{M})^{m}$. Now, we use the properties of the total derivative (see~\cite{berger1977nonlinearity} pages 69-70) to obtain
	%
	%
	%
 $
	\Vert D_{p}(\mathbf{S})\left\lbrace\boldsymbol{\xi}\right\rbrace\Vert\leq 
	\sum_{i=1}^{m}\left\Vert D_{p\vert\mathbf{S}_{i}}(\mathbf{S})\left\lbrace\boldsymbol{\xi}_{i}\right\rbrace\right\Vert
	,
 $
	%
	%
	%
	which leads to
	\begin{equation}
	\left\Vert p(\mathbf{S}+\boldsymbol{\xi})-
	p(\mathbf{S})\right\Vert\leq
	\sum_{i=1}^{m}\left\Vert D_{p\vert\mathbf{S}_{i}}(\mathbf{S})\left\lbrace\boldsymbol{\xi}_{i}\right\rbrace\right\Vert+\mathcal{O}\left(\Vert\boldsymbol{\xi}\Vert^{2}\right)
 .
	\end{equation}
    Then, taking into account that
	\begin{equation}
	\left\Vert p(\mathbf{S}+\boldsymbol{\xi})\mathbf{x}-
	p(\mathbf{S})\mathbf{x}\right\Vert\leq\Vert\mathbf{x}\Vert\left\Vert p(\mathbf{S}+\boldsymbol{\xi})-
	p(\mathbf{S})\right\Vert
	\end{equation}
	we have
		\begin{multline}
		\left\Vert 
		               p(\mathbf{S}+\boldsymbol{\xi})\bbx
		               -
		               p(\mathbf{S})\bbx
		\right\Vert
		\\
		\leq
		\left\Vert 
		       \bbx
		\right\Vert
		\left(
		\sum_{i=1}^{m}\left\Vert D_{p\vert\mathbf{S}_{i}}(\mathbf{S})\left\lbrace\boldsymbol{\xi}_{i}\right\rbrace\right\Vert+\mathcal{O}\left(\Vert\boldsymbol{\xi}\Vert^{2}\right)
		\right)
		,
		\end{multline}
	%
	%
	%
	
	and selecting $\boldsymbol{\xi}_{i}=\mathbf{T}(\mathbf{S}_{i})$ we complete the proof. 
\end{proof}


\subsection{Proof of  Theorem~\ref{theorem:uppboundDHmultg}}
\label{proof_theorem_DHvsDTmg}

\begin{proof}
	
	We start evaluating the expression $\bbD_{p\vert\bbS_{i}}$ in terms of the specific form of the perturbation to obtain
	\begin{multline}\label{eq_app_proof_theorem_DHvsDTmg0}
	\bbD_{p\vert\bbS_{}}(\bbS_1 , \ldots, \bbS_m)
	\left\lbrace 
	\bbT(\bbS_{i}) 
	\right\rbrace                  
	=
	\bbD_{p\vert\bbS_{i}}(\bbS_1,\ldots,\bbS_m)\left\lbrace \bbT_{i,0} \right\rbrace
	\\
	+
	\bbD_{p\vert\bbS_{i}}(\bbS_1,\ldots,\bbS_m)\left\lbrace \bbT_{i,1}\bbS_{i}\right\rbrace.
	\end{multline}
    Taking the norm in~(\ref{eq_app_proof_theorem_DHvsDTmg0}) and using the triangular inequality we have
	\begin{multline}\label{eq_app_proof_theorem_DHvsDTmg1}
	\left\Vert
	              \bbD_{p\vert\bbS_{i}}(\bbS_1,\ldots,\bbS_m)
	                  \left\lbrace 
	                         \bbT(\bbS_{i}) 
	                 \right\rbrace                  
	\right\Vert
	\leq 
	\\
	\left\Vert 
	\bbD_{p\vert\bbS_{i}}(\bbS_1,\ldots,\bbS_m)\left\lbrace \bbT_{i,0} \right\rbrace
	\right\Vert   
	\\
	+
	\left\Vert
	\bbD_{p\vert\bbS_{i}}(\bbS_1,\ldots,\bbS_m)\left\lbrace \bbT_{i,1}\bbS_{i}\right\rbrace
	\right\Vert. 
	\end{multline}
	Now we analyze each term in~(\ref{eq_app_proof_theorem_DHvsDTmg1}). First, we start  that the Fr\'echet derivative of $p(\bbS_1,\ldots,\bbS_m)$ with respect to $\bbS_i$ acting on $\bbT_{0,i}$ is a linear operator $\bbD_{p \vert\bbS_i}$ from $\End{M}$ into $\End{M}$ that can be expressed as
\begin{equation}
\bbD_{p \vert\bbS_i}
\{ \bbT_{0,i} \}
=
\sum_{r=1}^{\infty}
f_r (\bbS_1, \ldots, \bbS_m)
\bbT_{0,i}
h_{r}(\bbS_1, \ldots, \bbS_m)
,
\end{equation}
where $f_r (\bbS_1,\ldots,\bbS_m)$ and $h_r (\bbS_1,\ldots,\bbS_m)$ are polynomial functions of $\bbS_1,\ldots,\bbS_m$. As stated in~\cite{conway1994course} (p. 267), when the operator is assumed to be Hilbert-Schmidt (finite Frobenius norm) there is a unique linear operator 
$\overline{\bbD}_{ p \vert\bbS_i }(\bbS_1,\ldots,\bbS_m)
\left\lbrace 
\cdot
\right\rbrace \in \text{End}(\ccalM)^{\ast} \otimes \text{End}(\ccalM)$ with $\Vert \overline{\bbD}_{p\vert\mathbf{S}_1}(\bbS_1,\ldots,\bbS_m)
\left\lbrace 
\cdot
\right\rbrace\Vert = \Vert \bbD_{p\vert\mathbf{S}_1}(\bbS_1,\ldots,\bbS_m)\left\lbrace\cdot \right\rbrace \Vert $, acting on a vectorized version of $\bbT_{0,i}$. In particular we have
\begin{equation}\label{eq_frechet_dualotimes1c}
\overline{\bbD}_{ p \vert\bbS_i }(\bbS_1,\ldots,\bbS_m)
=
\sum_{r=1}^{\infty} 
h_{r}^{\ast}(\bbS_1,\ldots,\bbS_m)
\otimes
f_r (\bbS_1,\ldots,\bbS_m)
.
\end{equation}
Now, we recall that since the representation $(\ccalM,\rho)$ can be expressed as a direct sum of the irreducible subrepresentations. Then, any polynomial from $\ccalA$ mapped into $\End{M}$ by $\rho$ can be expressed in terms of such direct sum by means of  the spectral theorem representation (Theorem~\ref{thm_filtspec}), this is
\begin{equation}\label{eq_spec_dec_twogen1c}
f_r (\bbS_1,\ldots,\bbS_m)  
=
\sum_{k=1}^{q}f_r \left( 
\boldsymbol{\Lambda}_{1,k}, \ldots,\boldsymbol{\Lambda}_{m,k}
\right) 
\mathrsfso{P}_{k}    
,
\end{equation}
\begin{equation}\label{eq_spec_dec_twogen2c}
h_{r}^{\ast} (\bbS_1,\ldots,\bbS_m)  
=
\sum_{k=1}^{q}h_{r}^{\ast}\left( 
\boldsymbol{\Lambda}_{1,k}, \ldots, \boldsymbol{\Lambda}_{m,k}
\right) 
\mathrsfso{P}_{k}   
. 
\end{equation}
Then, substituting \eqref{eq_spec_dec_twogen1c} and \eqref{eq_spec_dec_twogen2c} in \eqref{eq_frechet_dualotimes1c} we have
\begin{multline}
\overline{\bbD}_{ p \vert\bbS_i }(\bbS_1,\ldots,\bbS_m)
=
\sum_{r=1}^{\infty}
\\
\left(
\sum_{k=1}^{q}h_{r}^{\ast}\left( 
\boldsymbol{\Lambda}_{1,k},  \ldots, \boldsymbol{\Lambda}_{m,k}
\right) 
\mathrsfso{P}_{k}
\otimes
\sum_{k=1}^{q}f_r \left( 
\boldsymbol{\Lambda}_{1,k}, \ldots, \boldsymbol{\Lambda}_{m,k}
\right) 
\mathrsfso{P}_{k}       
\right)
.
\end{multline}
Taking into account the properties of the tensor product it follows that:
%
%
%
\begin{multline}\label{eq_frechetonvecUi2c}
\overline{\bbD}_{ p \vert\bbS_i }(\bbS_1, \ldots, \bbS_m)
=
\sum_{k,\ell=1}^{q}
\\
\left(
\sum_{r=1}^{\infty}
\left(
h_{r}^{\ast}\left( 
\boldsymbol{\Lambda}_{1,k}, \ldots, \boldsymbol{\Lambda}_{m,k}
\right) 
\otimes 
f_r 
\left( 
\boldsymbol{\Lambda}_{1,\ell},  \ldots, \boldsymbol{\Lambda}_{m,\ell}
\right)            
\right) 
\left(                
\mathrsfso{P}_{k}
\otimes
\mathrsfso{P}_{\ell}  
\right)     
\right)
.
\end{multline}
where we have that
\begin{multline}
\sum_{r=1}^{\infty}
\left(
h_{r}^{\ast}\left( 
\boldsymbol{\Lambda}_{1,k}, \ldots, \boldsymbol{\Lambda}_{m,k}
\right) 
\otimes 
f_r 
\left( 
\boldsymbol{\Lambda}_{1,\ell},  \ldots, \boldsymbol{\Lambda}_{m,\ell}
\right)            
\right)  
\\
=
\left\lbrace 
\begin{matrix}
\overline{\bbD}_{p\vert \boldsymbol{\Lambda}_{i,k}} (\boldsymbol{\Lambda}_{1,k},\ldots,\boldsymbol{\Lambda}_{m,k})   &  \text{if}  &   k=\ell   \\
\boldsymbol{\Gamma}_{k,\ell}  &    \text{if}    & k\neq \ell
\end{matrix}
\right.
,
\end{multline}
where 
\begin{multline}
\overline{\bbD}_{p\vert \boldsymbol{\Lambda}_{i,k}} (\boldsymbol{\Lambda}_{1,k}, \ldots,\boldsymbol{\Lambda}_{m,k})
=
\\
\sum_{r=1}^{\infty}
\left(
h_{r}^{\ast}\left( 
\boldsymbol{\Lambda}_{1,k},\ldots, \boldsymbol{\Lambda}_{m,k}
\right) 
\otimes 
f_r 
\left( 
\boldsymbol{\Lambda}_{1,k}, \ldots, \boldsymbol{\Lambda}_{m,k}
\right)            
\right) 
,
\end{multline}
is uniquely associated to $\bbD_{p\vert \boldsymbol{\Lambda}_{i,k}} (\boldsymbol{\Lambda}_{1,k}, \ldots, \boldsymbol{\Lambda}_{m,k})$, which is the Fr\'echet derivative of $p\left( 
\boldsymbol{\Lambda}_{1,k}, \ldots,  \boldsymbol{\Lambda}_{m,k}
\right)  $ with respect o $\boldsymbol{\Lambda}_{i,k}$. Additionally, we point that the terms in the sum of~\eqref{eq_frechetonvecUi2c} are unique, and consequently can be expressed as a direct sum. This is,
\begin{multline}\label{eq_frechetonvecUi2d}
\overline{\bbD}_{ p \vert\bbS_i }(\bbS_1, \ldots,\bbS_m)
=
\bigoplus_{k,\ell=1}^{q}
\sum_{r=1}^{\infty}
\\
\left(
h_{r}^{\ast}\left( 
\boldsymbol{\Lambda}_{1,k}, \ldots, \boldsymbol{\Lambda}_{m,k}
\right) 
\otimes 
f_r 
\left( 
\boldsymbol{\Lambda}_{1,\ell},  \ldots, \boldsymbol{\Lambda}_{m,\ell}
\right)            
\right) 
\left(                
\mathrsfso{P}_{k}
\otimes
\mathrsfso{P}_{\ell}  
\right)     
.
\end{multline}
Then, if we calculate the norm of $\overline{\bbD}_{ p \vert\bbS_i }(\bbS_1, \ldots, \bbS_m)$ and we take into account the \textit{maximum property} (Definition~\ref{def_maxproperty}), it follows that
\begin{equation}
\left\Vert 
\overline{\bbD}_{ p \vert\bbS_i }(\bbS_1, \ldots, \bbS_m)
\right\Vert
=
\max_{(k,\ell)}
\left\lbrace 
\left\Vert
\overline{\bbD}_{p\vert \boldsymbol{\Lambda}_{i,k}}
\right\Vert
,
\Vert \boldsymbol{\Gamma}_{k,\ell} \Vert
\right\rbrace
.
\end{equation}
First, we recall that $\left\Vert
\overline{\bbD}_{p\vert \boldsymbol{\Lambda}_{i,k}}
\right\Vert = \left\Vert  \bbD_{p\vert \boldsymbol{\Lambda}_{i,k}}
\right\Vert$ and  since $p$ is $L_{0}$-Lipschitz by Theorem~\ref{theorem:FDLF} we have that $ \left\Vert  \overline{\bbD}_{p\vert \boldsymbol{\Lambda}_{i,k}}
\right\Vert \leq L_{0}$. We focus now on estimating the norm of $\boldsymbol{\Gamma}_{k\ell}$. First, we recall that
\begin{equation}
\boldsymbol{\Gamma}_{k\ell}
=
\sum_{r=1}^{\infty}
\left(
h_{r}^{\ast}\left( 
\boldsymbol{\Lambda}_{1,k}, \ldots,\boldsymbol{\Lambda}_{m,k}
\right) 
\otimes 
f_r 
\left( 
\boldsymbol{\Lambda}_{1,\ell}, \ldots, \boldsymbol{\Lambda}_{m,\ell}
\right)            
\right) 
,
\end{equation}
with $ k\neq \ell $. Now, we notice that it is possible to define a homomorphism $\theta$ between the algebras of operators $\text{End}(\ccalU_k)^{\ast} \otimes \text{End}(\ccalU_k )$ and $\text{End}(\ccalU_k)^{\ast} \otimes \text{End}(\ccalU_\ell )$ with $\ell<k$, which is given by
\begin{multline}
\theta  
\left( 
\sum_{r=1}^{\infty}
\left(
h_{r}^{\ast}\left( 
\boldsymbol{\Lambda}_{1,k}, \ldots, \boldsymbol{\Lambda}_{m,k}
\right) 
\otimes 
f_r 
\left( 
\boldsymbol{\Lambda}_{1,k},  \ldots, \boldsymbol{\Lambda}_{m,k}
\right)            
\right) 
\right)
\\        
=
\sum_{r=1}^{\infty}
\left(
h_{r}^{\ast}\left( 
\boldsymbol{\Lambda}_{1,k}, \ldots, \boldsymbol{\Lambda}_{m,k}
\right) 
\otimes 
f_r 
\left( 
\boldsymbol{\Lambda}_{1,\ell},  \ldots, \boldsymbol{\Lambda}_{m,\ell}
\right)            
\right) 
.
\end{multline}
By means of Theorem~\ref{theorem_normavsnormrhoa} it follows that
%
%
%
$
\left\Vert 
\boldsymbol{\Gamma}_{k\ell}  
\right\Vert
\leq 
\left\Vert
\overline{\bbD}_{p\vert \boldsymbol{\Lambda}_{i,k}}
\right\Vert 
,
$              
%
%
%
and therefore leads to
%
%
%
$
\left\Vert 
\boldsymbol{\Gamma}_{k\ell}  
\right\Vert
\leq 
L_0 
.    
$          
%
%
%
Then, with the calculations above we have
%
%
%
$
\left\Vert 
\bbD_{ p \vert\bbS_i }(\bbS_1, \ldots, \bbS_m)
\right\Vert
\leq
L_0 
.
$
%
%
%
We recall that for bounded operators we have
\begin{eqnarray*}
\left\Vert 
\bbD_{p\vert\bbS_{i}}(\bbS_1, \ldots, \bbS_m)\left\lbrace \bbT_{0,i} \right\rbrace
\right\Vert
\leq 
\left\Vert 
\bbD_{p\vert\bbS_{i}}(\bbS_1, \ldots, \bbS_m)\left\lbrace \bbT_{0,i} \right\rbrace
\right\Vert_{F} 
,             
\end{eqnarray*}
and as stated in~\cite{conway1994course} (p. 267) we also have that
\begin{eqnarray*}
\left\Vert 
\bbD_{p\vert\bbS_{i}}(\bbS_1, \dots, \bbS_m)\left\lbrace \bbT_{0,i} \right\rbrace
\right\Vert_{F}
\leq 
\left\Vert 
\bbD_{p\vert\bbS_{i}}(\bbS_1, \ldots, \bbS_m)
\right\Vert       
\left\Vert      
\bbT_{0,i} 
\right\Vert_{F} 
.             
\end{eqnarray*}
Then, taking into account that 
%
%
%
$
\left\Vert      
\bbT_{0,i} 
\right\Vert_{F} 
\leq 
\delta 
\left\Vert      
\bbT_{0,i} 
\right\Vert
,
$
%
%
%
we reach
\begin{equation}
\left\Vert 
\bbD_{p\vert\bbS_{i}}(\bbS_1, \ldots, \bbS_m)\left\lbrace \bbT_{0,i} \right\rbrace
\right\Vert
\leq 
L_0 \delta
\sup_{\bbS_i} \Vert 
\bbT(\bbS_i)
\Vert 
.             
\end{equation}

Now, we turn our attention to the term $ \left\Vert
\bbD_{p\vert\bbS_{i}}(\bbS_1, \ldots, \bbS_m)\left\lbrace \bbT_{1,i}\bbS_{i}\right\rbrace
\right\Vert  $ in~(\ref{eq_app_proof_theorem_DHvsDTmg1}).  We start using the notation
\begin{equation}
\bbD_{p\vert\bbS_{i}}(\bbS_1, \ldots, \bbS_m)\left\lbrace \bbT_{1,i}\bbS_{i}\right\rbrace
=
\widetilde{\bbD}_{p\vert\bbS_{i}}(\bbS_1, \ldots, \bbS_m)\left\lbrace \bbT_{1,i}\right\rbrace
.
\end{equation}
Now, we point out that
\begin{equation}
\widetilde{\bbD}_{p \vert\bbS_i}
\{ \bbT_{1,i}  \}
=
\sum_{r=1}^{\infty}
f_r (\bbS_1, \ldots,\bbS_m)
\bbT_{1,i}
h_{r}(\bbS_1, \ldots,\bbS_m)
,
\end{equation}
where $f_r (\bbS_1, \ldots, \bbS_m)$ and $h_r (\bbS_1, \ldots, \bbS_m)$ are polynomial functions of $\bbS_1, \ldots, \bbS_m$. When the operator is assumed to be Hilbert-Schmidt  there is a unique linear operator given by 
$\overline{\widetilde{\bbD}}_{ p \vert\bbS_i }(\bbS_1, \ldots, \bbS_m)
\left\lbrace 
\cdot
\right\rbrace \in \text{End}(\ccalM)^{\ast} \otimes \text{End}(\ccalM)$ with $\Vert \overline{\widetilde{\bbD}}_{p\vert\mathbf{S}_i}(\bbS_1, \ldots, \bbS_m)
\left\lbrace 
\cdot
\right\rbrace\Vert = \Vert \widetilde{\bbD}_{p\vert\mathbf{S}_i}(\bbS_1, \ldots,\bbS_m)\left\lbrace\cdot \right\rbrace \Vert $, acting on a vectorized version of $\bbT_{1,i}$~\cite{conway1994course} (p. 267). In particular we have
\begin{equation}\label{eq_frechet_dualotimes1b}
\overline{\widetilde{\bbD}}_{ p \vert\bbS_i }(\bbS_1, \ldots, \bbS_m)
=
\sum_{r=1}^{\infty} 
h_{r}^{\ast}(\bbS_1, \ldots, \bbS_m)
\otimes
f_r (\bbS_1, \ldots,\bbS_m)
.
\end{equation}
Since $(\ccalM,\rho)$ can be expressed as a direct sum of the irreducible subrepresentations. Then, any polynomial from $\ccalA$ mapped into $\End{M}$ by $\rho$ can be expressed in terms of such direct sum by means of  the spectral theorem representation (Theorem~\ref{thm_filtspec}), this is
\begin{equation}\label{eq_spec_dec_twogen1d}
f_r (\bbS_1, \ldots, \bbS_m)  
=
\sum_{k=1}^{q}f_r \left( 
\boldsymbol{\Lambda}_{1,k}, \ldots, \boldsymbol{\Lambda}_{m,k}
\right) 
\mathrsfso{P}_{k}    
,
\end{equation}
\begin{equation}\label{eq_spec_dec_twogen2d}
h_{r}^{\ast} (\bbS_1, \dots, \bbS_m)  
=
\sum_{k=1}^{q}h_{r}^{\ast}\left( 
\boldsymbol{\Lambda}_{1,k}, \ldots, \boldsymbol{\Lambda}_{m,k}
\right) 
\mathrsfso{P}_{k}   
. 
\end{equation}
Then, substituting \eqref{eq_spec_dec_twogen1d} and \eqref{eq_spec_dec_twogen2d} in \eqref{eq_frechet_dualotimes1b} we have
\begin{multline*}
\overline{\widetilde{\bbD}}_{ p \vert\bbS_i }(\bbS_1, \ldots, \bbS_m)
=
\sum_{r=1}^{\infty}
\\
\left(
\sum_{k=1}^{q}h_{r}^{\ast}\left( 
\boldsymbol{\Lambda}_{1,k},  \ldots, \boldsymbol{\Lambda}_{m,k}
\right) 
\mathrsfso{P}_{k}
\otimes
\sum_{k=1}^{q}f_r \left( 
\boldsymbol{\Lambda}_{1,k}, \ldots, \boldsymbol{\Lambda}_{m,k}
\right) 
\mathrsfso{P}_{k}       
\right)
.
\end{multline*}
Taking into account the properties of the tensor product it follows that:
%
%
%
\begin{multline}\label{eq_frechetonvecUi2e}
\overline{\widetilde{\bbD}}_{ p \vert\bbS_i }(\bbS_1, \ldots,\bbS_m)
=
\sum_{k,\ell=1}^{q}
\\
\left(
\sum_{r=1}^{\infty}
\left(
h_{r}^{\ast}\left( 
\boldsymbol{\Lambda}_{1,k}, \ldots,\boldsymbol{\Lambda}_{m,k}
\right) 
\otimes 
f_r 
\left( 
\boldsymbol{\Lambda}_{1,\ell}, \ldots,\boldsymbol{\Lambda}_{m,\ell}
\right)            
\right) 
\left(                
\mathrsfso{P}_{k}
\otimes
\mathrsfso{P}_{\ell}  
\right)     
\right)
,
\end{multline}
where we have that
\begin{multline}
\sum_{r=1}^{\infty}
\left(
h_{r}^{\ast}\left( 
\boldsymbol{\Lambda}_{1,k}, \ldots,\boldsymbol{\Lambda}_{m,k}
\right) 
\otimes 
f_r 
\left( 
\boldsymbol{\Lambda}_{1,\ell},  \ldots,\boldsymbol{\Lambda}_{m,\ell}
\right)            
\right)  
\\
=
\left\lbrace 
\begin{matrix}
\overline{\widetilde{\bbD}}_{p\vert \boldsymbol{\Lambda}_{i,k}} (\boldsymbol{\Lambda}_{1,k}, \ldots,\boldsymbol{\Lambda}_{m,k})   &  \text{if}  &   k=\ell   \\
\boldsymbol{\Gamma}_{k,\ell}  &    \text{if}    & k\neq \ell
\end{matrix}
\right.
,
\end{multline}
with
\begin{multline}
\overline{\widetilde{\bbD}}_{p\vert \boldsymbol{\Lambda}_{i,k}} (\boldsymbol{\Lambda}_{1,k}, \ldots,\boldsymbol{\Lambda}_{m,k})
\\
=
\sum_{r=1}^{\infty}
\left(
h_{r}^{\ast}\left( 
\boldsymbol{\Lambda}_{1,k}, \ldots,\boldsymbol{\Lambda}_{m,k}
\right) 
\otimes 
f_r 
\left( 
\boldsymbol{\Lambda}_{1,k},  \ldots, \boldsymbol{\Lambda}_{m,k}
\right)            
\right) 
,
\end{multline}
is uniquely associated to $\bbD_{p\vert \boldsymbol{\Lambda}_{i,k}} (\boldsymbol{\Lambda}_{1,k}, \ldots,\boldsymbol{\Lambda}_{m,k}) \{ (\cdot) \boldsymbol{\Lambda}_{i,k}  \}$. Like in the scenarios discussed above, we point that the terms in the sum of~\eqref{eq_frechetonvecUi2e} are unique, and consequently can be expressed as a direct sum as follows
\begin{multline}\label{eq_frechetonvecUi2otimesd}
\overline{\widetilde{\bbD}}_{ p \vert\bbS_i }(\bbS_1, \ldots, \bbS_m)
=
\bigoplus_{k,\ell=1}^{q}
\sum_{r=1}^{\infty}
\\
\left(
h_{r}^{\ast}\left( 
\boldsymbol{\Lambda}_{1,k}, \ldots,\boldsymbol{\Lambda}_{m,k}
\right) 
\otimes 
f_r 
\left( 
\boldsymbol{\Lambda}_{1,\ell}, \ldots,\boldsymbol{\Lambda}_{m,\ell}
\right)            
\right) 
\left(                
\mathrsfso{P}_{k}
\otimes
\mathrsfso{P}_{\ell}  
\right)     
.
\end{multline}
If we calculate the norm of $\overline{\widetilde{\bbD}}_{ p \vert\bbS_i }(\bbS_1, \ldots,\bbS_m)$ and we use the \textit{maximum property} (Definition~\ref{def_maxproperty}), it follows that
\begin{equation}
\left\Vert 
\overline{\widetilde{\bbD}}_{ p \vert\bbS_i }(\bbS_1, \ldots, \bbS_m)
\right\Vert
=
\max_{(k,\ell)}
\left\lbrace 
\left\Vert
\overline{\widetilde{\bbD}}_{p\vert \boldsymbol{\Lambda}_{i,k}}
\right\Vert
,
\Vert \boldsymbol{\Gamma}_{k,\ell} \Vert
\right\rbrace
.
\end{equation}
Since $\left\Vert
\overline{\widetilde{\bbD}}_{p\vert \boldsymbol{\Lambda}_{i,k}}
\right\Vert = \left\Vert  \widetilde{\bbD}_{p\vert \boldsymbol{\Lambda}_{i,k}}
\right\Vert$, and  $p$ is $L_{1}$-integral Lipschitz, we have that $ \left\Vert  \overline{\widetilde{\bbD}}_{p\vert \boldsymbol{\Lambda}_{i,k}}
\right\Vert \leq L_{1}$. Now we analyze the norm of $\boldsymbol{\Gamma}_{k\ell}$. First, we recall that
\begin{equation}
\boldsymbol{\Gamma}_{k\ell}
=
\sum_{r=1}^{\infty}
\left(
h_{r}^{\ast}\left( 
\boldsymbol{\Lambda}_{1,k}, \ldots,\boldsymbol{\Lambda}_{m,k}
\right) 
\otimes 
f_r 
\left( 
\boldsymbol{\Lambda}_{1,\ell}, \ldots,\boldsymbol{\Lambda}_{m,\ell}
\right)            
\right) 
,
\end{equation}
with $ k\neq \ell $. We point out that it is possible to define a homomorphism $\theta$ between the algebras of operators $\text{End}(\ccalU_k)^{\ast} \otimes \text{End}(\ccalU_k )$ and $\text{End}(\ccalU_k)^{\ast} \otimes \text{End}(\ccalU_\ell )$ with $k<\ell$, which is given by
\begin{multline}
\theta  
\left( 
\sum_{r=1}^{\infty}
\left(
h_{r}^{\ast}\left( 
\boldsymbol{\Lambda}_{1,k}, \ldots,\boldsymbol{\Lambda}_{m,k}
\right) 
\otimes 
f_r 
\left( 
\boldsymbol{\Lambda}_{1,k},  \ldots,\boldsymbol{\Lambda}_{m,k}
\right)            
\right) 
\right)
\\        
=
\sum_{r=1}^{\infty}
\left(
h_{r}^{\ast}\left( 
\boldsymbol{\Lambda}_{1,k}, \ldots,\boldsymbol{\Lambda}_{m,k}
\right) 
\otimes 
f_r 
\left( 
\boldsymbol{\Lambda}_{1,\ell},  \ldots,\boldsymbol{\Lambda}_{m,\ell}
\right)            
\right) 
.
\end{multline}
By Theorem~\ref{theorem_normavsnormrhoa}, it follows that
%
%
%
$
\left\Vert 
\boldsymbol{\Gamma}_{k\ell}  
\right\Vert
\leq 
\left\Vert
\overline{\widetilde{\bbD}}_{p\vert \boldsymbol{\Lambda}_{i,k}}
\right\Vert 
, 
$             
%
%
%
which leads to
%
%
%
$
\left\Vert 
\boldsymbol{\Gamma}_{k\ell}  
\right\Vert
\leq 
L_1
, 
$             
%
%
%
and therefore
%
%
%
$
\left\Vert 
\widetilde{\bbD}_{ p \vert\bbS_i }(\bbS_1, \ldots, \bbS_m)
\right\Vert
\leq
L_1
.
$
%
%
%
If $\widetilde{\bbD}_{ p \vert\bbS_i }(\bbS_1, \ldots,\bbS_m)$ is a bounded operator, we have
\begin{eqnarray*}
\left\Vert 
\widetilde{\bbD}_{p\vert\bbS_{i}}(\bbS_1, \ldots,\bbS_m)\left\lbrace \bbT_{1,i} \right\rbrace
\right\Vert
\leq 
\left\Vert 
\widetilde{\bbD}_{p\vert\bbS_{i}}(\bbS_1, \ldots,\bbS_m)\left\lbrace \bbT_{1,i} \right\rbrace
\right\Vert_{F} 
,             
\end{eqnarray*}
and as stated in~\cite{conway1994course} (p. 267) we also have that
\begin{eqnarray*}
\left\Vert 
\widetilde{\bbD}_{p\vert\bbS_{i}}(\bbS_1, \ldots,\bbS_m)\left\lbrace \bbT_{1,i} \right\rbrace
\right\Vert_{F}
\leq 
\left\Vert 
\widetilde{\bbD}_{p\vert\bbS_{i}}(\bbS_1, \ldots,\bbS_m)
\right\Vert       
\left\Vert      
\bbT_{1,i} 
\right\Vert_{F} 
.             
\end{eqnarray*}
Taking into account that 
%
%
%
$
\left\Vert      
\bbT_{1,i} 
\right\Vert_{F} 
\leq 
\delta 
\left\Vert      
\bbT_{1,i} 
\right\Vert
,
$
%
%
%
we reach
\begin{eqnarray}
\left\Vert 
\bbD_{p\vert\bbS_{i}}(\bbS_1, \ldots,\bbS_m)\left\lbrace \bbT_{1,i}\bbS_i \right\rbrace
\right\Vert
\leq 
L_1   
\delta
\left\Vert      
\bbT_{1,i} 
\right\Vert
,             
\end{eqnarray}
and since $\bbD_{\bbT}(\bbS_i)=\bbT_{1,i}$, this leads to
\begin{eqnarray}
\left\Vert 
\bbD_{p\vert\bbS_{i}}(\bbS_1, \ldots,\bbS_m)\left\lbrace \bbT_{1,i}\bbS_i \right\rbrace
\right\Vert
\leq 
L_1   
\delta
\sup_{\bbS_i}
\left\Vert      
\bbD_{\bbT}(\bbS_i)
\right\Vert
.             
\end{eqnarray}
\end{proof}

%% file: v25/appendix_HsFrechet.tex


\section{Preliminaries Irreducible Subrepresentations}
\label{prelim_irreduc_subrep}

In the following definition we define homomorphisms and isomorphisms of representations, which will allow us to state comparisons between representations.


\begin{definition}\label{def:homrep}
	Let $(\mathcal{M}_{1},\rho_{1})$ and $(\mathcal{M}_{2},\rho_{2})$ be two representations of an algebra $\mathcal{A}$. A homomorphism or interwining operator $\phi:\mathcal{M}_{1}\rightarrow\mathcal{M}_{2}$ is a linear operator which commutes with the action of $\mathcal{A}$ under $\rho_1$ and $\rho_2$, i.e.
	\begin{equation}
	\phi(\rho_{1}(a)v)=\rho_{2}(a)\phi(v)  \quad v\in\ccalM_1
	.
	\label{eq:interwinop}
	\end{equation}
	A homomorphism $\phi$ is said to be an isomorphism of representations if it is an isomorphism of vectors spaces. 
\end{definition}


It is worth pointing out that even when two representations have the same dimension, it might still happens that they are not homomorphic since the condition in~(\ref{eq:interwinop}) might not be satisfied. This is the typical scenario we have when considering representations of the polynomial algebra $\mbF [t]$, where we have irreducible representations that are one dimensional but in general non isomorphic.

Now, we highlight that irreducible representations are indecomposable -- the opposite is in general false -- and therefore, irreducible representations can be used to build unique decompositions. We focus our attention on the class of representations that can be decomposed as a direct sum of irreducible representations.


\section{Expanded example multigraphs}
\label{ex_mult_filt_extra}

\begin{remark}[Non commutative polynomial notation]\normalfont
Before we continue our presentation of non commutative signal models we make some remarks about the notation for non commutative polynomials. If a non commutative algebra $\ccalA$ is generated by $ \{ g_1, \ldots, g_m \} $, we define the set $\Pi_{j_1,\ldots,j_m}(g_1, \ldots, g_m)$ of normalized monomials built from products of the generators and where the term $g_i$ appears $j_i$ times. For instance, if the algebra has two generators $g_1$ and $g_2$ we have
\begin{equation}
\Pi_{1,2} (g_1, g_2) 
            = 
             \left\lbrace 
                          g_1 g_2^2 , g_2 g_1 g_2 , g_2^2 g_1
             \right\rbrace
             .
\end{equation}
Given an arbitrary ordering of the elements in $\Pi_{j_1, \ldots ,j_m}(g_1, \ldots, g_m)$ we will represent its $r$-th element by $\boldsymbol{\pi}_{j_1,\ldots ,j_m}^{r}(g_1, \ldots, g_m)$. Then, any polynomial in the non commutative algebra $\ccalA$ with $m$ generators can be written as
\begin{equation}
p(g_1, \ldots, g_m)
  =
   \sum_{j_1, \ldots ,j_m=0}\sum_{r}h_{j_1,\ldots,j_m ; r}\boldsymbol{\pi}_{j_1, \ldots ,j_m}^{r}(g_1, \ldots, g_m)
   ,
\end{equation}
where $h_{j_1,\ldots,j_m ; r}$ indicates the coefficients of the polynomial.     
\end{remark}


\begin{example}[Multigraph signal processing]\label{ex_multgsp}\normalfont

A multigraph consists of a common set of vertices and multiple separate sets of edges. Consider the specific case $G=\left( \ccalV, \{ \ccalE_1, \ccalE_2\} \right)$ consisting of a set of $N$ vertices $\ccalV$ and two separate edge sets $\ccalE_1$ and $\ccalE_2$ -- see Fig.~\ref{fig_multigraph}. Associated with each edge set we consider matrix representations $\bbS_1$ and $\bbS_2$. This is an algebraic signal processing model in which the vector space of signals $\ccalM=\mbC^{N}$ contains vectors in $\mbC^{N}$ with entries associated with each node of the graph and the algebra $\ccalA = \mbC[t_1,t_2]$ is the set of non commutative polynomials of two variables. Any polynomial in $ \mbC[t_1,t_2] $ can be written as
	\begin{equation}
	p(t_1, t_2) =
	              \sum_{j_1,j_2 = 0}^d\sum_{r}h_{j_1,j_2;r}\boldsymbol{\pi}_{j_1,j_2}^{r}(t_1, t_2)
	              .
	\end{equation}
The algebra $ \mbC[t_1,t_2] $ is generated by the monomials $t_1$ and $t_2$. The homomorphism $\rho$ is defined by mapping the generators $t_1$ and $t_2$ to $\rho(t_1)=\bbS_1$ and $\rho(t_2)=\bbS_2$, where the shift operators $\bbS_i$ are the matrix representations of the corresponding set of edges $\ccalE_i$. Then, the filtering in~\eqref{eqn_ASP_filter_outputs} takes the form
	\begin{multline}
	\rho \left(
	      \sum_{j_1,j_2 = 0}^d\sum_{r}h_{j_1,j_2;r}\boldsymbol{\pi}_{j_1,j_2}^{r}(t_1, t_2)
	     \right) 
	\bbx
	\\
	=
	\left(
	  \sum_{j_1,j_2 = 0}^d\sum_{r}h_{j_1,j_2;r}\boldsymbol{\pi}_{j_1,j_2}^{r}(\bbS_1, \bbS_2)
	\right)\bbx.   
	\end{multline}
\end{example}


\section{Proof of Theorem~\ref{thm_filtspec}}
\label{proof_thm_spec_filt}

Since $(\ccalM,\rho)$ can be written as a direct sum of irreducible subrepresentations $(\ccalU,\phi_i)$, we have that for any $a\in\ccalA$ the action of $\rho(a)$ on $\bbx \in\ccalM$ can be written as
$
\rho (a) 
=
\sum_{i=1}^{q}\phi_{i}(a)\mathrsfso{P}_{i}
.
$
Then, 
\begin{multline}
\rho(p(g_1,\ldots,g_m))
=
p \left( 
           \rho(g_1) , \ldots, \rho(g_m)
   \right)
=
\\   
p \left( 
            \sum_{i=1}^{q}\phi_{i}(g_1)\mathrsfso{P}_{i} , \ldots, \sum_{i=1}^{q}\phi_{i}(g_m)\mathrsfso{P}_{i}
   \right)
   .
\end{multline}
Since
\begin{equation}
\mathrsfso{P}_{j}\phi_i(a)\mathrsfso{P}_{i} 
=
\left\lbrace
\begin{matrix}
\phi_i(a)\mathrsfso{P}_{i}   &   &  i=j\\
0   &   &  i\neq j 
\end{matrix}
\right.
,
\end{equation}
it follows that
\begin{multline}
p \left( 
\sum_{i=1}^{q}\phi_{i}(g_1)\mathrsfso{P}_{i} , \ldots, \sum_{i=1}^{q}\phi_{i}(g_m)\mathrsfso{P}_{i}
\right)
\\
=
\sum_{i=1}^{q}p \left(
\boldsymbol{\Lambda}_{1,i}, \ldots ,\boldsymbol{\Lambda}_{m,i}
\right) 
\mathrsfso{P}_{i} 
,
\end{multline}
where $\boldsymbol{\Lambda}_{k,i} = \phi_i (g_k)$.


\section{Norms in Direct Sums of Spaces}
\label{app_normsofdirectsums}

In this section we discuss the properties of operator and matrix norms when working on direct sums of spaces. This plays a central role on how we calculate the norms of those operators associated to completely reducible representations. We start introducing the notion of absolute norm.


\begin{definition}[\cite{lancaster}]
	Let $\ccalX_1 , \ldots \ccalX_q$ be normed linear spaces and let $\ccalX$ be given by
	$
	     \ccalX = \ccalX_1 \oplus \ccalX_2 \oplus \ldots \oplus \ccalX_q
	     .
	$
	If $x,y\in\ccalX$ and $x = (x_1, x_2, \ldots , x_q)$, $y = (y_1, y_2, \ldots, y_q)$, we say that a norm on $\ccalX$ is absolute if $\Vert x_i\Vert = \Vert y_i \Vert$, $j=1,\ldots,n$, implies $\Vert x \Vert = \Vert y\Vert$. 
\end{definition}


Now, we introduce a property of norms of direct sums of spaces that will become relevant in our proofs.


\begin{definition}[\cite{lancaster}]\label{def_maxproperty}
A norm $\Vert \cdot\Vert$ on $ \ccalX =\bigoplus_{i=1}^{q}\ccalX_i $ has the maximum property if, for any linear operator $\bbT = \bigoplus_{i=1}^{q}\bbT_i$, we have that $\vertiii{\bbT} = \max_{i} \vertiii{\bbT_i }$, where $\vertiii{\bbT}$ indicates the operator norm induced by $\Vert \cdot\Vert$, and $\vertiii{\bbT_i}$ is the operator norm induced by the norm in $\ccalX_i$.
\end{definition}


Notice that the \textit{maximum property} is defined on $\Vert\cdot\Vert$ but measured or captured with $\vertiii{\cdot}$. The following results provides a formal connections between absolute norms and the maximum property.


\begin{theorem}[\cite{lancaster}]\label{thm_maxproperty}
	A norm on $\ccalX$ is absolute iff it has the maximum property.
\end{theorem}


As pointed out in~\cite{horn2012matrix} (p. 342) the Frobenius norm $\Vert\cdot\Vert_{F}$ is an absolute norm. This plays a central role when considering operators on completely reducible representations. To see this, we recall that if the representation $(\ccalM,\rho)$ -- of an algebra $\ccalA$ -- is completely reducible, it follows that $(\ccalM,\rho) \cong \bigoplus_{i=1}^{q}(\ccalU_i , \phi_i)$. Then, any operator $\rho (a)$ can be expressed as $\rho (a)= \bigoplus_{i=1}^{q}\phi_i (a)$ acting on $\bigoplus_{i=1}^{q}\ccalU_i$. Therefore, when computing the Frobenius norm of $\rho(a)$, we will have that $\vertiii{\rho(a)} = \max_{i} \vertiii{\phi_i (a)}$. Notice that this is consistent with the scenario of commutative algebras where the subpresentations are 1-dimensional and where $\phi_i$ is a scalar polynomial evaluated at the $i$th eigenvalue.


\section{Relationship between $\Vert a\Vert$ and $\Vert \rho(a)\Vert$}
\label{app_normavsnormrhoa}

\begin{theorem}\label{theorem_normavsnormrhoa}
	Let
	$\ccalA$ and $\ccalB$ be two polynomial algebras of operators endowed with norms satisfying that $\Vert aa^{\ast}\Vert = \Vert a\Vert^{2}$ and where $^{\ast}$ indicates the adjoint operation. Let $\theta$ be a homomorphism between $\ccalA$ and $\ccalB$ with $\theta(a^{\ast})=\theta(a)^{\ast}$. Then,	
%
%
%
$
\left\Vert
           \theta (a)
\right\Vert
           \leq 
\left\Vert
           a
\right\Vert.
$
%
%
%
\end{theorem}

\begin{proof}
First, notice that the unit element in $\ccalA$ given by $1_{\ccalA}$ is mapped into the unit element of $\ccalB$ given by $1_{\ccalB}$. Then, if $\mu\in\mathbb{C}$ is not in the spectrum of $a$, the term $\mu 1_{\ccalA}-a$ has an inverse $\left(\mu 1_{\ccalA}-a\right)^{-1}\in\ccalA$. Taking into account that $\theta\left( 1_{\ccalA}\right)=1_{\ccalB}$ we have that
%
%
%
$
\theta\left(
     \mu 1_{\ccalA}-a
\right)
\theta\left(
     \left(
     \mu 1_{\ccalA}-a
     \right)^{-1}
\right)
      =
\theta 
    \left(
          1_{\ccalA}
    \right)
           =
           1_{\ccalB}
           .
$
%
%
%
Then $\theta\left(\mu 1_{\ccalA}-a\right)= \mu 1_{\ccalB}-\theta(a)$ has an inverse in $\ccalA$ which implies that $\mu$ is not in the spectrum of $\theta(a)$. Consequently,
%
%
%
$
\text{spec}
     \left(
           \theta(a)
     \right) 
            \subset 
\text{spec}
     \left(
           a
     \right)   
     ,
$                 
%
%
%
where $\text{spec}
     \left(
           \theta(a)
     \right) $
and $\text{spec}
     \left(
           a
     \right) $ 
are the spectrum sets of $\theta(a)$ and $a$ respectively. Now, we take into account that
%
%
%
$
\left\Vert 
           a
\right\Vert^{2}
       =
\left\Vert 
           a a^{\ast}
\right\Vert  
       =
       r\left(
              a a^{\ast}
        \right)        
        , 
$        
%
%
%
$
\left\Vert 
           \theta(a)
\right\Vert^{2}
       =
\left\Vert 
           \theta(a) \theta(a)^{\ast}
\right\Vert  
       =
       r\left(
              \theta(a) \theta(a)^{\ast}
        \right)      
        ,  
$ 
%
%
where $r(\cdot)$ represents the spectral radius. Then, we have that  $r\left(\theta(a) \theta(a)^{\ast}\right) \leq r\left(a a^{\ast}\right) $ and therefore $\Vert \theta(a)\Vert \leq \Vert a\Vert$.
\end{proof}


\section{Fr\'echet Derivative of Lipschitz Functions}
\label{appendix_FDLF}


\begin{theorem}\label{theorem:FDLF}

Let $\bbF: \ccalV \to \ccalU$ a Lipschitz map between two Banach spaces $\ccalV$ and $\ccalU$ with Lipschitz constant $L_{0}$. Let $\bbD_{\bbF\vert\bbS}(\bbS)$ be the Fr\'echet derivative of $\bbF$ at $\bbS\in\ccalV$. Then, if $\bbD_{\bbF\vert\bbS}(\bbS)$ exists we have that
%
%
%
$
\left\Vert
                \bbD_{\bbF\vert\bbS}(\bbS)
\right\Vert
                \leq 
                       L_{0}
                       .
$                       
%
%
%
\end{theorem}

\begin{proof}
	
We start taking into account the definition of the Fr\'echet derivative of $\bbF(\bbS)$, 
\begin{equation}\label{eq_frechet_littleo}
\bbD_{\bbF\vert\bbS} \{ \boldsymbol{\xi} \}
         =
         \bbF (\bbS +\boldsymbol{\xi}) 
         -
         \bbF (\bbS)
         -
         o(\boldsymbol{\xi})
         .
\end{equation}
Taking the norm on both sides of \eqref{eq_frechet_littleo} and applying the triangular inequality this leads to
\begin{equation}
\left\Vert
\bbD_{\bbF\vert\bbS} \{ \boldsymbol{\xi} \}
\right\Vert
\leq
\left\Vert
\bbF (\bbS +\boldsymbol{\xi}) 
-
\bbF (\bbS)
\right\Vert
+
\left\Vert
o(\boldsymbol{\xi})
\right\Vert
.
\end{equation}
Since $\bbF(\bbS)$ is $L_0$-Lipschitz, it follows  that
\begin{equation}
\left\Vert
\bbD_{\bbF\vert\bbS} \{ \boldsymbol{\xi} \}
\right\Vert
\leq
\left\Vert
\boldsymbol{\xi}
\right\Vert
\left(
L_0
+
\frac{
\left\Vert
o(\boldsymbol{\xi})
\right\Vert
}
{
	\Vert \boldsymbol{\xi} \Vert
	}
\right)
.
\end{equation}
Now, we recall one of the definitions of the operator norm~\cite{conway1994course} (p. 12)
we have
%
%
%
$
\Vert 
       \bbD_{\bbF\vert\bbS} 
\Vert   
         \leq 
               L_0
               +
               \frac{
               	\left\Vert
               	o(\boldsymbol{\xi})
               	\right\Vert
               }
               {
               	\Vert \boldsymbol{\xi} \Vert
               }
               .
$               
%
%
%
Finally, taking into account that $
	\left\Vert
	o(\boldsymbol{\xi})
	\right\Vert
/
	\Vert \boldsymbol{\xi} \Vert
\rightarrow 0$ as $\Vert \boldsymbol{\xi}\Vert\rightarrow 0$, it follows that
%
%
%
$
\Vert 
\bbD_{\bbF\vert\bbS} 
\Vert   
\leq 
L_0
.
$
\end{proof}

%
%
\subsection{Proof of Theorems~\ref{theorem:stabilityAlgNN0} and~\ref{theorem:stabilityAlgNN1}}


\subsubsection{Proof of Theorem~\ref{theorem:stabilityAlgNN0}}
\label{prooftheorem:stabilityAlgNN0}

\begin{proof}
	Considering~(\ref{eq:xl}), applying the operator norm, and taking into account that the maps $\sigma_{\ell}$ are Lipschitz with constant $C_{\ell}$, we have 
	\begin{multline}
	\left\Vert 
	\sigma_{\ell}
	\left(
	p(\bbS_{1,\ell}, \ldots,\bbS_{m,\ell})
	\mathbf{x}_{\ell-1}\right)
	-
	\sigma_{\ell}
	\left(
	p(\tilde{\mathbf{S}}_{1,\ell}, \ldots,\tilde{\bbS}_{m,\ell})
	\mathbf{x}_{\ell-1}
	\right)
	\right\Vert
	\\
	\leq
	C_{\ell}\boldsymbol{\Delta}_{\ell}
	\Vert
	\mathbf{x}_{\ell-1}
	\Vert
	,
	\end{multline}
	where $\boldsymbol{\Delta}_{\ell}=\Vert p(\mathbf{S}_{1,\ell}, \ldots,\bbS_{m,\ell})-p(\tilde{\mathbf{S}}_{1,\ell}, \ldots, \tilde{\bbS}_{m,\ell})\Vert$, and whose value is determined by theorems~\ref{theorem:HvsFrechetmultgen} and~\ref{theorem:uppboundDHmultg}.
\end{proof}


\subsubsection{Proof of Theorem~\ref{theorem:stabilityAlgNN1}}
\label{prooftheorem:stabilityAlgNN1}

We start writing the norm of the difference between the $f$-th feature and its perturbed version in the $\ell$-th layer. 

\begin{proof}

	\begin{multline}
	\left\Vert
	\mathbf{x}_{\ell}^{f}
	-
	\tilde{\mathbf{x}}_{\ell}^{f}
	\right\Vert
	\leq
	\left\Vert
	\sigma_{\ell-1}\sum_{g_{\ell-1}}
	\rho_{\ell-1}\left( \xi^{g_\ell g_{\ell-1}} \right)
	\sigma_{\ell-2}\right.\\
	\left.\sum_{g_{\ell-2}}
	\rho_{\ell-2}\left( \xi^{g_{\ell-1}g_{\ell-2}} \right)
	\cdots\sigma_{1}\sum_{g_{1}}
	\rho_{1}\left( \xi^{g_{2}g_{1}}\right)
	\mathbf{x}\right.
	-
	\\
	\left. \sigma_{\ell-1}\sum_{g_{\ell-1}}
	\tilde{\rho}_{\ell-1}\left( \xi^{g_\ell g_{\ell-1}} \right)
	\sigma_{\ell-2}
	\sum_{g_{\ell-2}}
	\tilde{\rho}_{\ell-2}\left( \xi^{g_{\ell-1}g_{\ell-2}} \right)
	\right.
	\\
	\left.\cdots\sigma_{1}\sum_{g_{1}}
	\tilde{\rho}_{1}\left( \xi^{g_{2}g_{1}}\right)
	\mathbf{x}
	\right\Vert.
	\label{eq:longalgNNexp1}
	\end{multline}
	To  expand and simplify~(\ref{eq:longalgNNexp1}), we point that if $A_{\ell+1}$ is the filter in the layer $\ell+1$, we have 
	\begin{multline}
	A_{\ell+1}\sigma_{\ell}(a)-\tilde{A}_{\ell+1}\sigma_{\ell}(\tilde{a})
	=
	\\
	(A_{\ell+1}-\tilde{A}_{\ell+1})\sigma_{\ell}(a)+\tilde{A}_{\ell+1}(\sigma_{\ell}(a)-\sigma_{\ell}(\tilde{a})),
	\end{multline}
	where the tilde indicates perturbation. Additionally, we take into account that $\Vert\sigma_{\ell}(a)-\sigma_{\ell}(b)\Vert\leq C_{\ell}\Vert a-b\Vert$, $\Vert A_{\ell+1}-\tilde{A}_{\ell+1}\Vert\leq\boldsymbol{\Delta}_{\ell}$ and $\Vert A_{\ell+1}\Vert\leq B_{\ell+1}$. We synthesize these facts to write:
	\begin{equation}
	\sum_{g_{k}}
	\Vert
	\alpha-\tilde{\alpha}
	\Vert
	\leq
	\sum_{g_{k}}
	\left( 
	\boldsymbol{\Delta}_{k}\Vert\sigma_{k-1}(\alpha)\Vert+B_{k}C_{k-1}\Vert\beta-\tilde{\beta}\Vert
	\right)
	,
	\label{eq:aux_longeq1}
	\end{equation}
	\begin{equation}
	\sum_{g_{k}}
	\Vert
	\beta-\tilde{\beta}
	\Vert
	\leq
	\sum_{g_{k}}\sum_{g_{k-1}}\Vert\alpha-\tilde{\alpha}\Vert
	,
	\label{eq:aux_longeq2}
	\end{equation}
	\begin{equation}
	\sum_{g_{k}}\Vert\sigma_{k-1}(\alpha)\Vert\leq\left(\prod_{r=1}^{k}F_{r}\right)\left(\prod_{r=1}^{k-1}C_{r}B_{r}\right)\Vert\mathbf{x}\Vert
	,
	\label{eq:aux_longeq3}
	\end{equation}
	where $\alpha$ and $\tilde{\alpha}$ represent sequences of terms in~(\ref{eq:longalgNNexp1}) that start with a symbol of the type $\rho_{\ell}$, and $\beta$ and $\tilde{\beta}$ indicate a sequence of symbols that start with a summation symbol. The tilde makes reference to symbols that are associated to the perturbed representations. 
	
	The symbol $\boldsymbol{\Delta}_{\ell}$ is the difference between the operators and their perturbed versions (see Definition~\ref{def:stabilityoperators1}) in the layer $\ell$  whose values are given according to Theorems~\ref{theorem:HvsFrechetmultgen} and~\ref{theorem:uppboundDHmultg}. Putting~(\ref{eq:aux_longeq1}),~(\ref{eq:aux_longeq2}) and~(\ref{eq:aux_longeq3}) together we have:
	\begin{multline}\label{eq:aux1}
	\left\Vert
	\mathbf{x}_{L}^{f}-\tilde{\mathbf{x}}_{L}^{f}
	\right\Vert
	\leq
	\sum_{\ell=1}^{L}\boldsymbol{\Delta}_{\ell}\left(\prod_{r=\ell}^{L}C_{r}\right)\left(\prod_{r=\ell+1}^{L}B_{r}\right)\\
	\left(\prod_{r=\ell}^{L-1}F_{r}\right)\left(\prod_{r=1}^{\ell-1}C_{r}F_{r}B_{r}\right)\left\Vert\mathbf{x}\right\Vert,
	\end{multline}
	where the products $\prod_{r=a}^{b}F(r)=0$ if $b<a$. With~(\ref{eq:aux1}) at hand we take into account that
	\begin{multline}
	\left\Vert
	\Phi\left(\mathbf{x},\{ \mathcal{P}_{\ell} \}_{1}^{L},\{ \mathcal{S}_{\ell}\}_{1}^{L}\right)-
	\Phi\left(\mathbf{x},\{ \mathcal{P}_{\ell} \}_{1}^{L},\{ \tilde{\mathcal{S}}_{\ell}\}_{1}^{L}\right)
	\right\Vert^{2}
	\\
	=
	\sum_{f=1}^{F_{L}}\left\Vert\mathbf{x}_{L}^{f}-\tilde{\mathbf{x}}_{L}^{f}
	\right\Vert^{2}
	,
	\end{multline}
	which leads to
	\begin{multline}
	\left\Vert
	\Phi\left(\mathbf{x},\{ \mathcal{P}_{\ell} \}_{1}^{L},\{ \mathcal{S}_{\ell}\}_{1}^{L}\right)-
	\Phi\left(\mathbf{x},\{ \mathcal{P}_{\ell} \}_{1}^{L},\{ \tilde{\mathcal{S}}_{\ell}\}_{1}^{L}\right)
	\right\Vert
	\\
	\leq
	\sqrt{F_{L}}
	\sum_{\ell=1}^{L}\boldsymbol{\Delta}_{\ell}\left(\prod_{r=\ell}^{L}C_{r}\right)\left(\prod_{r=\ell+1}^{L}B_{r}\right)
	\left(\prod_{r=\ell}^{L-1}F_{r}\right)
	\\
	\left(\prod_{r=1}^{\ell-1}C_{r}F_{r}B_{r}\right)\left\Vert\mathbf{x}\right\Vert
	.
	\end{multline}
\end{proof}


\section{Spectral Representations: Additional Material}
\label{sec_spect_rep_multiplicty}

A general formulation of the Fourier decomposition in an algebraic signal model considers subrepresentations that can be isomorphic to a direct sum of other irreducible subrepresentations themselves. This can be derived from the following theorem.


\begin{theorem}(\cite{repthysmbook})\label{thm_decompirreduciblerep1}
	  Let $(\ccalU_i , \rho_i)$, $1 \leq i \leq m$, be irreducible finite dimensional
	   pairwise nonisomorphic representations of $\ccalA$, and let $(\ccalW, \theta)$ be
	  a subrepresentation of $\left( \ccalM = \bigoplus_{i=1}^{m} n_i \ccalU_i, \rho\right)$. Then, $(\ccalW, \theta)$ is isomorphic to $\left( \bigoplus_{i=1}^{m}r_i \ccalU_i , \right)$, $r_i \leq n_i$ and the inclusion 
	  $\phi: \ccalW \longrightarrow \ccalV$ is a direct sum of inclusions $\phi_i : r_i \ccalU_i \longrightarrow n_i \ccalU_i$.
\end{theorem} 


Theorem~\ref{thm_decompirreduciblerep1} has two implications. First, it states that if a given representation $(\ccalV,\rho)$ is expressed as a direct sum of irreducible representations, then any subrepresentation of $(\ccalV,\rho)$ is isomorphic to a sum of such irreducible subrepresentations. Second, it indicates that in order to provide a complete description of a given representation it might be necessary to use more than once a given irreducible representation. The direct sum of all irreducible subrepresentations that are isomorphic is traditionally called the \textit{homogeneous component} of a representation. This is if $(\mathcal{U}_i,\rho_i)$ is an irreducible subrepresentation of $(\ccalM,\rho)$ which is a representation of $\ccalA$, we denote by $\ccalM(\mathcal{U}_i)=\bigoplus_{(\ccalU,\theta)\cong(\mathcal{U}_i,\rho_i)} (\ccalU,\theta)$ the homogeneous component associated to $(\mathcal{U}_i,\rho_i)$. Since $\ccalM(\mathcal{U}_i)$ is a direct sum, it has a length denoted by $m(\ccalU_i,\ccalM)$ and is called the \textit{multiplicity} of $(\mathcal{U}_i,\rho_i)$ in $(\ccalM,\rho)$~\cite{repthybigbook}. 

As pointed out in~\cite{repthybigbook} if a representation $(\ccalM,\rho)$ is completely reducible, it is possible to show that $(\ccalM,\rho)\cong \text{soc}\{ \ccalM \}$, where $\text{soc}\{ \ccalM \}$ is the direct sum of all irreducible subrepresentations of $(\ccalM,\rho)$. Additionally, it also happens that $\text{soc}\{ \ccalM \}=\bigoplus_{\ccalU\sim\text{Irr}\{ \ccalA\}} \ccalM(\ccalU)$ where $\text{Irr}\{ \ccalA \}$ indicates the class of irreducible representations of $\ccalA$.

With these notions at hand, the most general formulation of the Fourier decomposition in an algebraic signal model is given in the following definition.


\begin{definition}[Fourier Decomposition]\label{def:foudecomp_multiplicity}
For an algebraic signal model $(\mathcal{A},\mathcal{M},\rho)$ we say that there is a spectral or Fourier decomposition if
\begin{equation}
(\mathcal{M},\rho)\cong\bigoplus_{(\mathcal{U}_{i},\phi_{i})\in\text{Irr}\{\mathcal{A}\}}(\mathcal{U}_{i},\phi_{i})^{\oplus m(\mathcal{U}_{i},\mathcal{M})}
,
 \label{eq:foudecomp1_multiplicity}
\end{equation}
where the $(\mathcal{U}_{i},\phi_{i})$ are irreducible subrepresentations of $(\mathcal{M},\rho)$. Any signal $\mathbf{x}\in\mathcal{M}$ can be therefore represented by the map,
\begin{equation}
\Delta: \mathcal{M} \to \bigoplus_{(\mathcal{U}_{i},\phi_{i})\in\text{Irr}\{\mathcal{A}\}}\mathcal{U}_{i}^{\oplus m(\mathcal{U}_{i},\mathcal{M})} 
\label{eq:foudecomp2_multiplicity}
\end{equation}
\begin{equation*}
\mathbf{x}\mapsto \hat{\mathbf{x}}
,
\end{equation*}
known as the Fourier decomposition of $\mathbf{x}$ and the projection of $\hat{\mathbf{x}}$ in each $\mathcal{U}_{i}$ are the Fourier components represented by $\hat{\mathbf{x}}(i)$.
\end{definition}




\subsection{Computational Aspects and Examples}
\label{sec_spect_rep_calculations}

The numerical calculation to find decompositions in terms of irreducible subrepresentations of an algebra entails numerical challenges equivalent to those of finding joint block diagonalizations~\cite{lux2010representations,draexler2012computational}. Indeed, for a signal model $(\ccalA,\ccalM,\rho)$ finding a decomposition of $\ccalM$ in terms of irreducibles is equivalent to building a joint diagonalization of the shift operators when $\text{dim}(\ccalM)<\infty$ and $\ccalA$ has a finite number of generators~\cite{barot2014introduction}. If on addition to this the algebra is commutative, it is possible to show that the irreducibles are one dimensional~\cite{repthybigbook,repthysmbook} and therefore finding irreducible subrepresentations is equivalent to find eigenspaces or eigenvector decompositions.


\begin{example}\normalfont
In this example we show how the Fourier decomposition in GSP is a particular case of the Fourier decomposition expressed in~\ref{def:foudecomp_multiplicity}. Let us recall that the algebraic signal model of the GSP framework is given by $(\mbC [t], \mbR^N,\rho)$ where $\mbC [t]$ is the algebra of polynomials of indepenent variable $t$, $N=\vert \ccalV \vert$ is the number nodes in the graph and $\rho(t) = \bbS$ where $\bbS$ is the matrix representation of the underlying graph. 

Since $\mbC [t]$ is commutative, the irreducible subrepresentations of $(\mbR^N,\rho)$ are given by $(\text{span}(\bbv_i),\phi_i)$ where $\phi_i (t) = \lambda_i$ with $a\in\ccalA$, $\bbv_i$ and $\lambda_i$ are the $i$-th eigenvector and the $i$-th eigenvalue of $\bbS_i$, respectively. Then, following Definition~\ref{def:foudecomp_multiplicity} we have
\begin{equation}
(\mbR^N,\rho)
       \cong
           \bigoplus_{i=1}^{N}(\text{span}(\bbv_i),\phi_i)^{\oplus m\left(\text{span}(\bbv_i), \mbR^N\right)}
,
\end{equation}
where $m\left(\text{span}(\bbv_i), \mbR^N\right)$ indicates the multiplicity of the $i$-th eigenvalue. Notice that for any $p(t)\in \mbC [t]$ we have that $ \rho (p(t)) = p(\bbS)$ and $\phi_i (p(t)) = \phi_i (\lambda_i)$.
\end{example}
